\documentclass{article}

\pdftrailerid{}

\usepackage{colm2024_conference}

\usepackage{graphicx}
\usepackage{float}
\usepackage{caption}
\usepackage[utf8]{inputenc} %
\usepackage[T1]{fontenc}    %
\usepackage{hyperref}       %
\usepackage{url}            %
\usepackage{booktabs}       %
\usepackage{amsfonts}       %
\usepackage{nicefrac}       %
\usepackage{microtype}      %
\usepackage{placeins}
\usepackage{xcolor}     %
\usepackage{colortbl}
\usepackage{wrapfig}
\usepackage{marvosym}

\makeatletter
\newcommand\blfootnote[1]{%
  \begingroup
  \renewcommand\thefootnote{}\footnote{#1}%
  \addtocounter{footnote}{-1}%
  \endgroup
}
\makeatother

\definecolor{darkred}{RGB}{177, 38, 26}
\definecolor{darkblue}{RGB}{67, 116, 177}
\definecolor{darkgreen}{rgb}{0.0, 0.5, 0.0}

\definecolor{bestcol}{RGB}{  0,102,204} %
\definecolor{goodcol}{RGB}{ 34,139, 34} %
\definecolor{deltaBg}{RGB}{220,230,255} %
\newcommand{\best}[1]{\textbf{\textcolor{bestcol}{#1}}}

\usepackage[inline]{enumitem}

\usepackage[most]{tcolorbox}
\usepackage{listings}
\usepackage{etoc}

\tcbset{
  colback=white, %
  colframe=black, %
  arc=2mm,            %
  boxrule=1pt,      %
  left=6pt, right=6pt, top=6pt, bottom=6pt  %
}

\usepackage{epigraph}
\usepackage{amsmath, amssymb}
\usepackage[capitalize,noabbrev]{cleveref}

\usepackage{multirow}
\usepackage{array} 
\usepackage{tabularx}
\usetikzlibrary{calc}

\newlist{inlinelist}{enumerate*}{1}
\setlist[inlinelist]{label=(\roman*)}

\newcommand{\CiteParen}[1]{\citep{#1}}

\makeatletter
\renewenvironment{abstract}{\vskip.075in\centerline{\large\bf Abstract}\vspace{0.5ex}\begin{quote}}{\par\end{quote}\vskip 1ex}
\makeatother

\title{HopChain: Multi-Hop Data Synthesis for Generalizable Vision-Language Reasoning}

\author{%
 \textbf{Shenzhi Wang$^{1,2,\ast}$, Shixuan Liu$^{1,\ast}$, Jing Zhou$^{1}$, Chang Gao$^{1}$, Xiong-Hui Chen$^{1}$, Binghai Wang$^{1}$, An Yang$^{1}$, Shiji Song$^{2}$, Bowen Yu$^{1,\textrm{\Letter},\dagger}$,  Gao Huang$^{2,\textrm{\Letter}}$, Junyang Lin$^{1}$} \\
 $^1$ Qwen Team, Alibaba Inc. \qquad $^2$ LeapLab, Tsinghua University
}

\begin{document}

\blfootnote{\hspace*{-1.5em}$^\ast$Equal contribution \quad $\textrm{\Letter}$: Corresponding authors \quad $\dagger$: Project lead}
\blfootnote{\hspace*{-1.5em}\mbox{\footnotesize Emails: \texttt{wangshenzhi99@gmail.com}, \texttt{liushixuan66@gmail.com}, \texttt{yubowen.ph@gmail.com}, \texttt{gaohuang@tsinghua.edu.cn}}}

\maketitle

\begin{abstract}
  Vision-language models (VLMs) show strong multimodal capabilities but still struggle with fine-grained vision-language reasoning.
We find that long chain-of-thought (CoT) reasoning exposes diverse failure modes, including perception, reasoning, knowledge, and hallucination errors, which can compound across intermediate steps.
However, most existing vision-language data used for reinforcement learning with verifiable rewards (RLVR) does not involve complex reasoning chains that rely on visual evidence throughout, leaving these weaknesses largely unexposed.
We therefore propose \textbf{HopChain}, a scalable framework for synthesizing \emph{multi-hop vision-language reasoning} data for RLVR training of VLMs.
Each synthesized multi-hop query forms a logically dependent chain of instance-grounded hops, where earlier hops establish the instances, sets, or conditions needed for later hops, while the final answer remains a specific, unambiguous number suitable for verifiable rewards.
We train Qwen3.5-35B-A3B and Qwen3.5-397B-A17B under two RLVR settings: the original data alone, and the original data plus HopChain's multi-hop data, and compare them across 24 benchmarks spanning STEM and Puzzle, General VQA, Text Recognition and Document Understanding, and Video Understanding.
Although this multi-hop data is not synthesized for any specific benchmark, it improves 20 of 24 benchmarks on both models, indicating broad and generalizable gains.
Consistently, replacing full chained queries with half-multi-hop or single-hop variants reduces the average score across five representative benchmarks from 70.4 to 66.7 and 64.3, respectively.
Notably, multi-hop gains peak in long-CoT vision-language reasoning, exceeding 50 points in the ultra-long-CoT regime.
These experiments establish HopChain as an effective, scalable framework for synthesizing multi-hop data that improves generalizable vision-language reasoning.

\end{abstract}

\section{Introduction} \label{sec: introduction}

\begin{figure}[!t]
  \centering
  \includegraphics[width=1\textwidth]{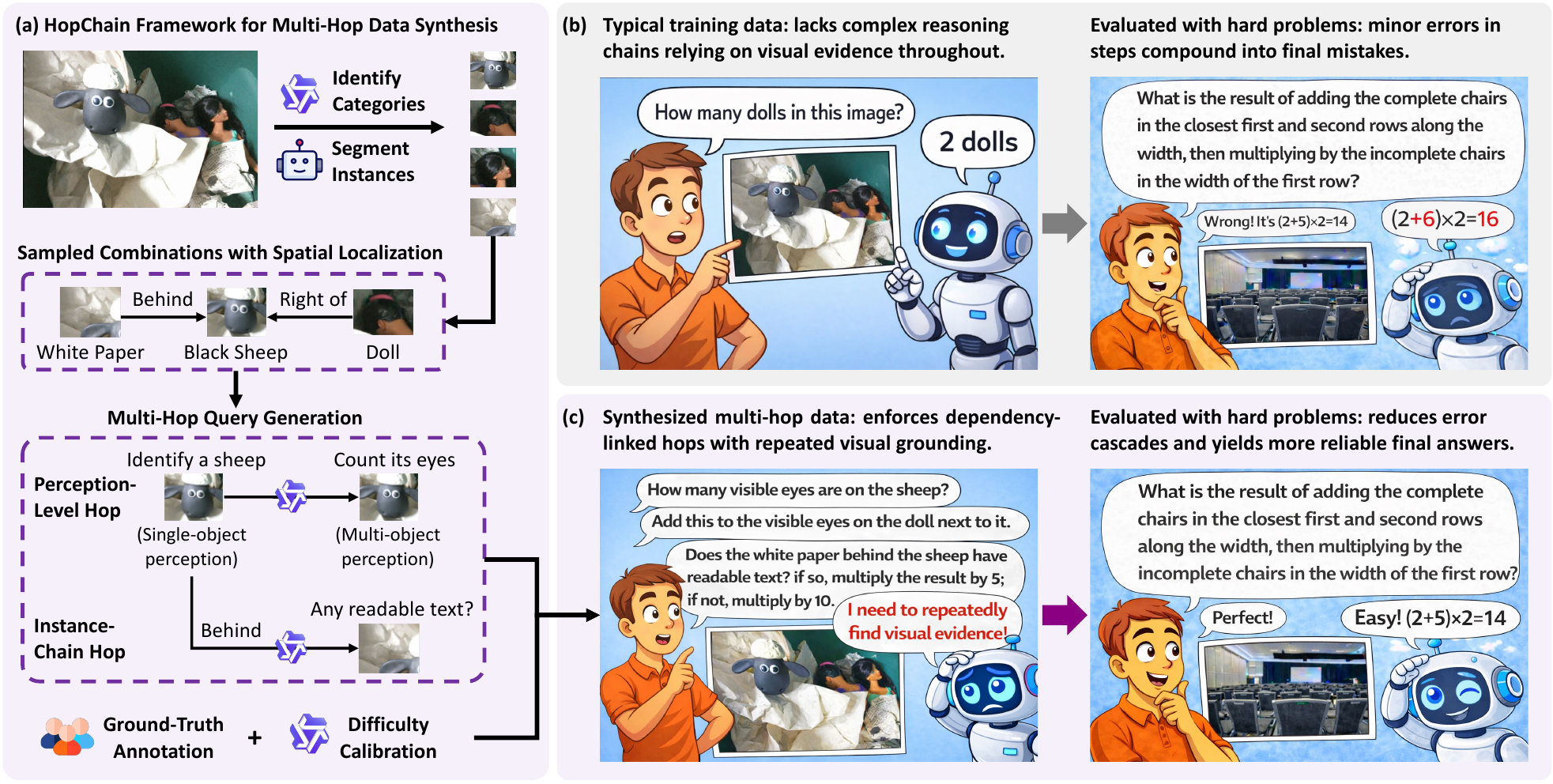}
  \caption{Overview of HopChain and the motivation for multi-hop vision-language reasoning data. \textbf{(a)} HopChain synthesizes multi-hop data through four stages: category identification, instance segmentation, multi-hop query generation, and ground-truth annotation with difficulty calibration. \textbf{(b)} Typical vision-language training data often does not require complex reasoning chains that rely on visual evidence throughout; on hard problems, minor mistakes introduced at intermediate long-CoT steps can compound into final errors. \textbf{(c)} In contrast, the multi-hop data synthesized by HopChain forms logically dependent hop chains in which later hops depend on instances, sets, or conditions established by earlier hops, while nearly every hop requires fresh visual re-grounding. This training signal encourages continual visual evidence seeking during long-CoT reasoning, improving robustness and reducing compounding errors.}
  \label{fig:example_image}
  \vspace{-6pt}
\end{figure}

Vision-language models (VLMs) have achieved impressive performance across a diverse set of multimodal benchmarks by integrating visual encoders with large language models~\CiteParen{liu2024visual,liu2024improved,bai2023qwenvl,bai2025qwen25vl,chen2024internvl,openai2023gpt4,geminiteam2024gemini}.
Recent advances in reinforcement learning with verifiable rewards (RLVR) have further improved VLMs' reasoning abilities by training them to produce step-by-step chain-of-thought solutions for questions with objectively verifiable answers~\CiteParen{shao2024deepseekmath,deepseekr1}.
Despite this progress, a critical gap remains: VLMs frequently struggle with tasks that require \emph{fine-grained, multi-step vision-language reasoning}, where the correct answer depends on carefully attending to multiple visual elements and their relationships within an image~\CiteParen{bigverdi2025perception,ye2025blinktwice,jiang2025vlmr3}.

A natural question arises: \emph{what prevents current VLMs from performing robust vision-language reasoning?}
In \cref{sec: unreliable visual perception in long-chain reasoning}, we analyze a central challenge:
VLMs exhibit \textbf{diverse and compounding failure modes during long chain-of-thought (CoT) reasoning}, as longer reasoning chains may cause models to attend less faithfully to the image, execute incorrect intermediate reasoning steps, rely on incomplete knowledge, or produce hallucinated and weakly grounded intermediate content that then compounds across subsequent steps.
Related failure patterns have also been reported in prior analyses of amplified multimodal hallucination, evidential drift, object hallucination, visual illusion, and image-context reasoning errors~\CiteParen{liu2025morethinking,luo2025thinkingdrifts,rohrbach2018object,guan2024hallusionbench,leng2024vcd}.
Critically, as depicted in \cref{fig:example_image}(b), most existing vision-language training data does not involve particularly complex reasoning chains that depend on visual evidence throughout the process.
As a result, these long-CoT weaknesses remain largely unexposed during training.
This observation suggests that relying on, or simply expanding, existing vision-language RLVR training data is insufficient; what is needed is training data that \emph{structurally forces} the model to seek visual evidence at each step of long-CoT reasoning, strengthening step-by-step vision-language reasoning and improving generalization across diverse scenarios.

Motivated by this insight, we propose \textbf{HopChain}\footnote{Throughout this paper, HopChain refers to our synthesis framework, while \emph{multi-hop vision-language reasoning data}, or simply \emph{multi-hop data}, refers to the training data synthesized by HopChain.}, a scalable framework for synthesizing multi-hop vision-language reasoning data specifically for RLVR training of VLMs.
Each synthesized multi-hop query forms a logically dependent chain of instance-grounded hops, where earlier hops establish the instances, sets, or conditions needed for later hops, forcing repeated visual re-grounding throughout training rather than language-only shortcuts.
At the same time, each query terminates in a specific, unambiguous numerical answer that is easy to verify for RLVR; moreover, because the hops are logically dependent, obtaining the correct final number usually also requires the intermediate reasoning steps to be correct.
Rather than targeting a narrow synthetic task, we use the multi-hop data synthesized by HopChain as a complementary source of RLVR training data to strengthen fundamental vision-language reasoning under long CoT and support broad, generalizable gains across diverse domains.
Importantly, this synthesized data is not tailored to any particular downstream benchmark; instead, it is constructed to strengthen general-purpose vision-language reasoning, which we later show transfers broadly across benchmark families.
We define this multi-hop structure through two hop types: perception-level hops and instance-chain hops.
A \emph{perception-level hop} switches between single-object perception (e.g., read text, identify color, determine position) and multi-object relationship reasoning (e.g., compare sizes, count objects satisfying a condition, determine spatial arrangement), while remaining grounded in the instances, sets, or conditions established by earlier hops.
An \emph{instance-chain hop} follows an explicit dependency chain (e.g., instance A $\rightarrow$ B $\rightarrow$ C), where the next instance can be identified only from the instances, sets, or conditions established by earlier hops.
We require each query to combine both hop types in a logically dependent chain, where earlier hops establish the instances, sets, or conditions needed for later hops.
This design forces fresh visual grounding at every step, blocks language-only shortcuts, and exposes diverse long-CoT failure modes during training, as shown in \cref{fig:example_image}(c).

To construct such multi-hop data at scale, HopChain adopts a scalable four-stage data synthesis pipeline:
(1)~\emph{category identification} via Qwen3-VL-235B-A22B-Thinking~\CiteParen{qwen3vl2025} to enumerate semantic categories in each image;
(2)~\emph{instance segmentation} via SAM3~\CiteParen{carion2025sam3} to localize individual instances for the identified semantic categories;
(3)~\emph{multi-hop query generation} via Qwen3-VL-235B-A22B-Thinking that constructs logically chained questions over combinations of instances; and
(4)~\emph{human-in-the-loop verification} where multiple annotators independently solve each query, and only queries with the same final numerical answer are retained as valid training examples.
The pipeline provides a scalable data synthesis workflow, scaling to broad image collections with sufficient detectable instances while maintaining strict quality control, as summarized in \cref{fig:example_image}(a).

We apply RLVR with Soft Adaptive Policy Optimization (SAPO)~\CiteParen{sapo} on the multi-hop data synthesized by HopChain to train VLMs.
We validate the effectiveness of the multi-hop data synthesized by HopChain on Qwen3.5-35B-A3B and Qwen3.5-397B-A17B~\CiteParen{qwen35blog} across 24 benchmarks spanning STEM and Puzzle, General VQA, Text Recognition and Document Understanding, and Video Understanding.
Compared with RLVR on the original RLVR data alone, adding this multi-hop data improves 20 out of 24 benchmarks on both Qwen3.5-35B-A3B and Qwen3.5-397B-A17B, despite the fact that the synthesized data is not designed for any specific benchmark, indicating broad and generalizable performance gains across diverse scenarios.
To demonstrate that full chained queries are important, we replace them with half-multi-hop or single-hop variants on Qwen3.5-35B-A3B, reducing the average score across five representative benchmarks from 70.4 to 66.7 and 64.3, respectively.
Multi-hop training substantially strengthens long-CoT vision-language reasoning, with gains peaking at more than 50 accuracy points in the ultra-long-CoT regime on Qwen3.5-397B-A17B.
When we independently sample each synthesized query multiple times on both Qwen3.5-35B-A3B and Qwen3.5-397B-A17B, more than half of the queries are partially solved and the outcomes are distributed relatively evenly from fully incorrect to fully correct, indicating a broad difficulty range suitable for both smaller and larger models. The corrected error types also closely follow the original error distribution before adding this multi-hop data, indicating gains that are broad rather than confined to a single narrow failure mode.
These extensive experiments establish HopChain as an effective framework for synthesizing multi-hop data that improves generalizable vision-language reasoning capabilities beyond the synthesized training distribution.

\vspace{0.3em}
\noindent Our main contributions are as follows:
\begin{itemize}[leftmargin=1.5em, itemsep=2pt, topsep=2pt]
    \item We identify diverse and compounding failure modes during long CoT reasoning as a key barrier to vision-language generalization, and show why relying on or simply expanding existing vision-language RLVR training data is insufficient.
    \item We formalize multi-hop vision-language reasoning with perception-level and instance-chain hops, and build HopChain, a scalable synthesis pipeline whose queries form logically dependent chains in which earlier hops establish the instances, sets, or conditions needed for later hops, forcing repeated visual re-grounding throughout training while keeping the final answer directly verifiable for RLVR.
    \item Through extensive experiments, we verify that RLVR on the multi-hop data synthesized by HopChain yields broad, generalizable gains: 20 out of 24 benchmarks improve on both Qwen3.5-35B-A3B and Qwen3.5-397B-A17B, while additional analyses show that preserving full chained queries is important, multi-hop data strengthens long-CoT vision-language reasoning, and the synthesized data spans a broad difficulty range while enabling trained models to correct a broad range of errors.
\end{itemize}

\section{Preliminaries}

\subsection{Reinforcement Learning with Verifiable Rewards for Vision-Language Models}

The reinforcement learning with verifiable rewards (RLVR) framework for vision-language models (VLMs) closely parallels that for large language models (LLMs), with the primary distinction being that RLVR for VLMs processes both an image and a text query as input to generate a textual chain-of-thought culminating in a verifiable answer prediction, whereas RLVR for LLMs operates solely on text queries.
Specifically, RLVR for VLMs aims to maximize the following objective:
\begin{align}
    &J(\pi) = \mathbb{E}_{(\boldsymbol{I}, \boldsymbol{q}, \boldsymbol{a}) \sim \mathcal{D}, \boldsymbol{o} \sim \pi(\cdot \mid \boldsymbol{I}, \boldsymbol{q})}[R(\boldsymbol{o}, \boldsymbol{a})],
    \\
    \text{where} & \quad  R(\boldsymbol{o}, \boldsymbol{a}) = \begin{cases} 1.0 & \text{if}\ \texttt{is\_equivalent}(\boldsymbol{o}, \boldsymbol{a}), \\ 0.0 & \text{otherwise.} \end{cases} \label{eq: reward calculation}
\end{align}
Here, $\boldsymbol{I}$, $\boldsymbol{q}$, and $\boldsymbol{a}$ denote the image, text query, and ground-truth answer, respectively, sampled from dataset $\mathcal{D}$, and $\boldsymbol{o}$ represents the response generated by policy $\pi$ conditioned on $\boldsymbol{I}$ and $\boldsymbol{q}$.

\subsection{Soft Adaptive Policy Optimization}

Soft Adaptive Policy Optimization (SAPO, \CiteParen{sapo}) is introduced to mitigate the potential instability and inefficiency caused by hard clipping in prior RLVR algorithms such as GSPO~\CiteParen{gspo} and GRPO~\CiteParen{shao2024deepseekmath}.
Concretely, SAPO substitutes hard clipping with a temperature-controlled soft gate and optimizes the following objective for VLMs:
\begin{equation}
\mathcal{J}(\theta)
=
\mathbb{E}_{(\boldsymbol{I}, \boldsymbol{q}, \boldsymbol{a}) \sim \mathcal{D}, \{\boldsymbol{o}_i\}_{i=1}^G \sim \pi_{\text{old}}(\cdot \mid \boldsymbol{I}, \boldsymbol{q})}
\left[
\frac{1}{G}
\sum_{i=1}^{G}
\frac{1}{|\boldsymbol{o}_i|}
\sum_{t=1}^{|\boldsymbol{o}_i|}
f_{i,t}\!\left(r_{i,t}(\theta)\right)
\hat{A}_{i,t}
\right],
\end{equation}
\begin{align}
\text{where} \quad &r_{i,t}(\theta)
=
\frac{
\pi_{\theta}\!\left(\boldsymbol{o}_{i,t} \mid \boldsymbol{I}, \boldsymbol{q}, \boldsymbol{o}_{i,<t}\right)
}{
\pi_{\theta_{\text{old}}}\!\left(\boldsymbol{o}_{i,t} \mid \boldsymbol{I}, \boldsymbol{q}, \boldsymbol{o}_{i,<t}\right)
},
\qquad
\hat{A}_{i,t}
=
\hat{A}_i
=
\frac{
R_i - \operatorname{mean}\!\left(\{R_j\}_{j=1}^{G}\right)
}{
\operatorname{std}\!\left(\{R_j\}_{j=1}^{G}\right)
},
\\
&f_{i,t}(x)
=
\sigma\!\left(\tau_{i,t}(x - 1)\right)
\cdot
\frac{4}{\tau_{i,t}},
\qquad
\tau_{i,t}
=
\begin{cases}
\tau_{\text{pos}}, & \text{if } \hat{A}_{i,t} > 0, \\
\tau_{\text{neg}}, & \text{otherwise}.
\end{cases}
\end{align}
Here, $i, j$ denote the sample indices in $\mathcal{D}$, and $t$ is the token index within a sequence. 
The parameters of the currently trained policy and the old rollout policy are denoted by $\theta$ and $\theta_{\text{old}}$, respectively. 
The temperatures for positive and negative tokens are $\tau_{\text{pos}}$ and $\tau_{\text{neg}}$, respectively. 
Moreover, $R_i$ is computed according to \cref{eq: reward calculation} for the $i$-th sample, and $\sigma(x) = \frac{1}{1 + e^{-x}}$ denotes the sigmoid function.

\section{Diverse Failure Modes in Long Chain-of-Thought Reasoning} \label{sec: unreliable visual perception in long-chain reasoning}

\begin{figure}[t]
    \centering
    \includegraphics[pagebox=cropbox,width=0.8\textwidth]{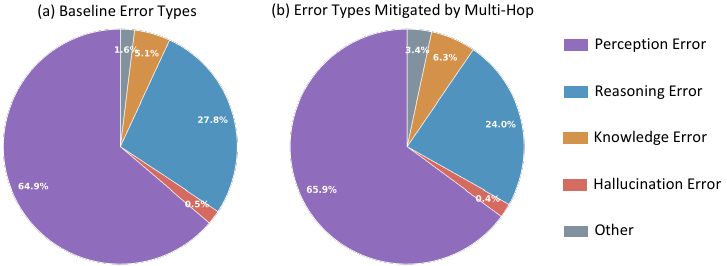}
    \caption{Error-type distributions before and after adding multi-hop data in RLVR. Subfigure (a) shows the error-type distribution of \texttt{RLVR w/o Multi-Hop} on the benchmarks in Tables~\ref{tab:exp-35b} and~\ref{tab:exp-397b}. Subfigure (b) shows the distribution of baseline error types mitigated after adding multi-hop data (i.e., cases corrected by \texttt{RLVR w/ Multi-Hop}). In (a), errors are diverse, with perception and reasoning errors as the dominant categories. In (b), the distribution remains similar to (a), indicating that multi-hop data mitigates failures in a generalizable way rather than improving only a narrow error type. \Cref{fig:qualitative_examples} provides representative perception reasoning failures, together with responses corrected after adding multi-hop data.}
    \label{fig:error_distribution}
\end{figure}

Long CoT reasoning places a much higher demand on visual grounding than short, single-step QA.
To answer correctly, a VLM must repeatedly return to the image, recover the right object, attribute, relation, or numeric evidence at each step, and use that evidence to determine the next step of reasoning.
This makes long CoT reasoning inherently fragile: multiple error types can emerge along the chain, including perception, reasoning, knowledge, and hallucination errors, among others.
Once any such error appears at an intermediate step, the remaining reasoning can still look coherent while operating on flawed intermediate evidence, eventually producing an incorrect final answer.
Recent analyses in multimodal reasoning report similar patterns: longer reasoning traces can drift away from image-grounded evidence, reduce attention to visual inputs, and amplify hallucinated intermediate content~\CiteParen{liu2025morethinking,luo2025thinkingdrifts}.
More generally, diagnostic studies of VLMs have also documented persistent object hallucination and visual illusion under image-context reasoning~\CiteParen{rohrbach2018object,guan2024hallusionbench}.

\begin{figure}[!h]
    \centering
    \includegraphics[width=1\textwidth]{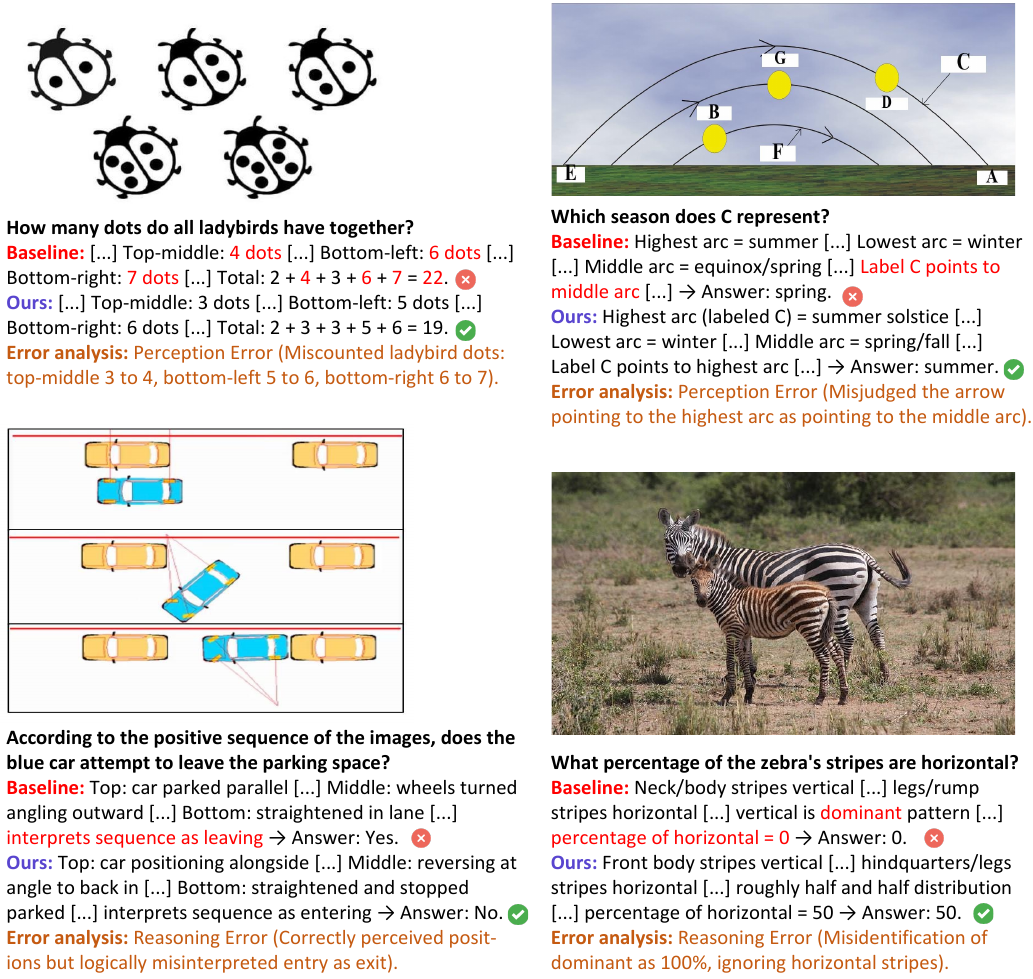}
    \caption{Qualitative examples of unreliable visual perception in long CoT reasoning. We show failure cases from the benchmarks in \cref{tab:exp-35b,tab:exp-397b} using ``RVLR w/o Multi-Hop'' (baseline), alongside correct answers from ``RVLR w/ Multi-Hop'' (ours). For brevity, unimportant parts of the long CoT are omitted with $[\dots]$. The baseline error is highlighted in red, and the failure reason is given in the error analysis.}
    \label{fig:qualitative_examples}
    \vspace{-5pt}
\end{figure}

To obtain the error breakdown in \cref{fig:error_distribution}, we analyze responses produced on the benchmarks in \cref{tab:exp-397b} by Qwen3.5-397B-A17B under the \texttt{RLVR w/o Multi-Hop} setting. For each benchmark, we randomly sample 20 incorrect responses and ask annotators to identify the primary failure type by comparing the model output against the ground-truth answer.
The resulting breakdown shows that long-CoT failures are diverse rather than concentrated in a single category: perception errors are the largest group, while reasoning, knowledge, and hallucination errors are also present in the breakdown.
Qualitative examples in \cref{fig:qualitative_examples} further show that these failures are not limited to a narrow benchmark type.
The baseline model (\texttt{RLVR w/o Multi-Hop}) miscounts small local details in the ladybird example, misjudges the contact relation between the gripper and the dress strap, misreads the sign shape in the driving scene, reads chart values incorrectly, follows the wrong arc in the astronomy diagram, and selects the wrong body part in the fish illustration.
These examples cover natural images, charts, and scientific diagrams, but they share the same structure: one faulty intermediate step appears in the middle of a long reasoning chain, and the later reasoning steps inherit that mistake.
In this sense, long-CoT errors are often coupled rather than isolated: a mistaken visual judgment can trigger faulty reasoning, unsupported inference, or other downstream failures.
By contrast, \texttt{RLVR w/ Multi-Hop} is more likely to recover the correct visual evidence at each hop and therefore reaches the correct final prediction.

This analysis suggests that the central challenge is not merely producing a longer textual CoT, but maintaining reliable, step-wise reasoning over visual evidence throughout the chain across diverse visual scenarios.
This interpretation is also aligned with recent work arguing that multimodal reasoning is often bottlenecked by perception quality and can benefit from stronger intermediate perception, repeated image-grounded observation, or iterative revisiting of visual regions~\CiteParen{bigverdi2025perception,ye2025blinktwice,jiang2025vlmr3}.
Consequently, improving long CoT reasoning requires an RLVR data construction method that is applicable to diverse images, forces repeated grounding during reasoning, and trains the model to use the result of one step to locate, verify, or constrain the next.
This is exactly the goal of the HopChain framework introduced in the next section.

\section{Boosting Vision-Language Generalization by Synthesizing Multi-Hop Data}
\label{sec: method}

\begin{figure}[!h]
    \centering
    \includegraphics[width=0.97\textwidth]{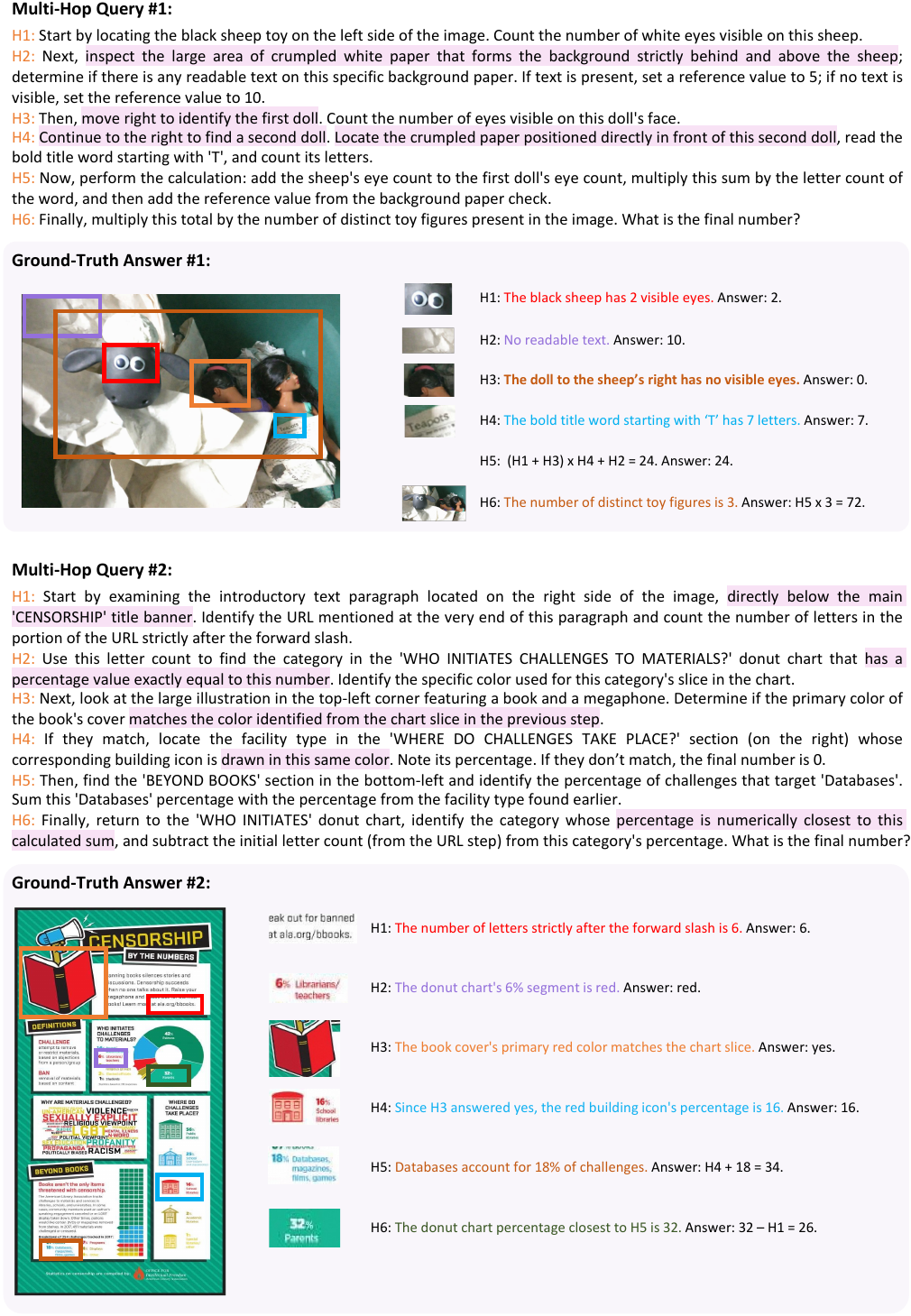}
    \caption{Examples of synthesized multi-hop data. In each query, purple-highlighted text denotes the instance chain, meaning that the instance required at the current hop can only be identified from instances established in earlier hops. In the corresponding image, we mark the instance regions involved in each hop with colored rectangles, and the rectangle colors are aligned with the text colors of the corresponding hop-wise ground-truth answers.}
    \label{fig:multihop-query-examples}
    \vspace{-15pt}
\end{figure}

The analysis in \cref{sec: unreliable visual perception in long-chain reasoning} shows that long-CoT reasoning does not fail for a single reason: models may make perception, reasoning, knowledge, hallucination, and other errors, and these errors often propagate once an incorrect intermediate step is carried forward through the rest of the chain.
This motivates training data beyond simply relying on, or expanding, existing vision-language RLVR training data. Specifically, the desired training data should structurally force the model to seek visual evidence at each step of long-CoT reasoning, thereby strengthening step-by-step vision-language reasoning and improving generalization across diverse scenarios.
Accordingly, we synthesize multi-hop vision-language reasoning data whose hops are designed to instantiate exactly this requirement, while ensuring that each query terminates in a specific, unambiguous numerical answer compatible with RLVR, as illustrated in \cref{fig:multihop-query-examples}.

\subsection{Multi-Hop Vision-Language Reasoning Definition}
\label{subsec:mh-vl-definition}

We first define the structure of the target multi-hop queries. To do so, we use three reasoning levels: Levels~1 and~2 describe what an individual reasoning step asks the model to do, while Level~3 describes a query that chains multiple Level~1 and Level~2 steps together.
\textbf{Level~1} is single-object perception, such as reading text or identifying object attributes, including color, shape, size, position, and category.
\textbf{Level~2} is multi-object perception, such as spatial, comparative, or counting relations across objects.
\textbf{Level~3} is multi-hop reasoning, where multiple Level~1 and Level~2 steps are chained into one query.

Within such a Level~3 query, consecutive hops can be linked in two complementary dimensions.
\textbf{Perception-level hop:} the next step changes the kind of perception being performed, for example from a Level~1 single-object judgment to a Level~2 relational judgment, or vice versa, while remaining grounded in the instances, sets, or conditions established by earlier hops.
\textbf{Instance-chain hop:} the next step moves to a new instance along an explicit dependency chain (e.g., instance A $\rightarrow$ B $\rightarrow$ C), where the next instance can only be identified from the instances, sets, or conditions established by earlier hops.

Each query must satisfy three structural conditions: (\romannumeral1) it must be Level~3, (\romannumeral2) it must combine both hop types, and (\romannumeral3) its hops must form a logically dependent chain in which earlier hops establish the instances, sets, or conditions needed for later hops.
We further prefer instance dependency chains and perception-level transitions to be intertwined as tightly as possible.
In addition, because the synthesized data is intended for RLVR, each query must terminate in a specific, unambiguous numerical answer.
This makes the final answer easy to verify for RLVR, while the logical dependence among hops means that obtaining the correct final number usually also requires the intermediate reasoning chain to be correct.
This definition is intended to rule out pseudo-multi-hop queries in which substeps are loosely connected or can be bypassed with shallow shortcuts.
Intuitively, it targets exactly the kind of failures shown in \cref{fig:qualitative_examples}: the model must repeatedly identify the correct visual evidence, then use that evidence to determine the next object, relation, or computation.

\subsection{Data Synthesis Pipeline}
\label{subsec:data-synthesis-pipeline}

Given the query definition in Section~\ref{subsec:mh-vl-definition}, we next describe how candidate multi-hop queries are synthesized. We build the dataset through a three-stage synthesis pipeline that operationalizes this design goal into trainable RLVR data (see \cref{fig:example_image}(a) for the full four-stage workflow, where this subsection covers Stages~1--3 and \cref{subsec:ground-truth-annotation} covers Stage~4).
The pipeline is designed so that the final queries are not only complex, but also structurally aligned with the long-CoT setting we care about: multiple intermediate hops, each grounded in visual evidence and required for later hops.

\textbf{Stage 1: Category Identification.}
We first identify the candidate visual entities that later hops can operate over.
Given an input image, we use a VLM (Qwen3-VL-235B-A22B-Thinking) to identify semantic categories present in the image, yielding a list of semantic categories (e.g., ``car,'' ``person,'' ``sign'') without localization.

\textbf{Stage 2: Instance Segmentation.}
We then resolve these categories into concrete instances, because hop-based reasoning must be anchored to particular objects rather than generic labels.
For each identified semantic category, we use SAM3 to generate segmentation masks and bounding boxes for candidate instances of that category where possible, producing a set of individual instances with spatial localization.

\textbf{Stage 3: Multi-Hop Query Generation.}
This stage is where we convert detected instances into training questions that enforce chained visual reasoning rather than isolated recognition.
We form combinations of 3--6 instances and use Qwen3-VL-235B-A22B-Thinking to generate multi-hop queries for each combination.
The model receives the original image plus cropped patches of each instance, where the patches are used only at design time to help identify the appearance and location of each instance. The full prompt used for multi-hop query synthesis is provided in Appendix~\ref{app:multihop-prompt}.
Crucially, to address the failure modes identified in \cref{sec: unreliable visual perception in long-chain reasoning}, we impose several constraints on each synthesized query: it must (i) involve as many instances as possible from the selected combination while remaining answerable from the original image alone; (ii) describe objects only by spatial, contextual, or visual attributes; (iii) terminate in a specific, unambiguous numerical answer; and (iv) avoid any reference to segmentation masks, bounding boxes, or patch images.
These constraints prevent the synthesis process from leaking shortcuts that would let the model solve the query from auxiliary cues unavailable at training or test time.
In addition, reasoning hops must form a logically dependent chain: each hop must either change the perception level or move along an instance dependency chain, while earlier hops establish the instances, sets, or conditions needed for later hops.
As a result, the synthesized queries better mirror the structure of long-CoT reasoning: the model must recover and retain intermediate visual evidence and use it to determine the next step, rather than rely on an early guess or a language-only heuristic.

\subsection{Ground-Truth Annotation and Difficulty Calibration}
\label{subsec:ground-truth-annotation}

After the synthesis pipeline in Section~\ref{subsec:data-synthesis-pipeline} produces candidate queries, we introduce a separate Stage~4 with two complementary components: \emph{ground-truth annotation} and \emph{difficulty calibration}.
In this stage, ground-truth annotation filters out low-quality queries and provides reliable numerical ground-truth answers for the retained ones, while difficulty calibration uses these verified answers to remove queries that are too easy.

\textbf{Stage 4: Ground-Truth Annotation and Difficulty Calibration.}
For each generated query, four annotators from our data annotation team independently solve it.
During ground-truth annotation, we discard queries where annotators report ambiguous references or other quality issues, and for queries that pass, we take the shared final numerical answer (all four must agree) as the ground truth.
For difficulty calibration, we evaluate the retained queries on a weaker model with eight sampled responses per query, using the verified ground-truth answers as reference; queries on which the weaker model achieves $100\%$ accuracy are removed as too easy, and the remaining queries form our final dataset with verified difficulty and reliable ground-truth annotations.

Together, the pipeline functions as a scalable data synthesis workflow: from raw images to verified multi-hop queries, it can be applied to broad image collections with sufficient detectable instances, while maintaining quality control through human and model-based verification.
In this way, the synthesized data is not merely harder data.
It is constructed to expose the model to the same kind of chained evidence use that long-CoT reasoning requires in practice, thereby providing a more suitable training signal for improving robustness and generalization across diverse visual scenarios.
A direct comparison between the typical vision-language reasoning question in \cref{fig:qualitative_examples} and the synthesized multi-hop data in \cref{fig:multihop-query-examples} makes this difference concrete: simply scaling existing RLVR-style data often still leaves shallow or loosely coupled solution paths, whereas our synthesized data enforces hop-wise dependency and repeated visual re-grounding throughout the chain.
This structural gap is precisely why relying on, or merely expanding, existing vision-language RLVR training data is insufficient.

\section{Experiments}
\label{sec:experiments}

\newcommand{\MethodOne}{\shortstack[c]{Before RLVR}}
\newcommand{\MethodTwo}{\shortstack[c]{RLVR w/o Multi-Hop}}
\newcommand{\MethodThree}{\shortstack[c]{RLVR w/ Multi-Hop}}
\newcolumntype{Y}{>{\centering\arraybackslash}X}
\newlength{\BenchmarkColWidth}
\newlength{\ResultColWidth}
\setlength{\BenchmarkColWidth}{0.18\linewidth}
\setlength{\ResultColWidth}{\dimexpr(\linewidth-\BenchmarkColWidth)/3\relax}
\newcommand{\ExperimentHeaderRow}{%
\multicolumn{4}{@{}l@{}}{%
  \begingroup
  \setlength{\fboxsep}{0pt}%
  \colorbox{black!5}{%
    \parbox{\linewidth}{%
      \strut
      \makebox[\BenchmarkColWidth][l]{Benchmark}%
      \makebox[\ResultColWidth][c]{\MethodOne}%
      \makebox[\ResultColWidth][c]{\MethodTwo}%
      \makebox[\ResultColWidth][c]{\MethodThree}%
      \strut
    }%
  }%
  \endgroup
}%
\\}

We evaluate HopChain from three complementary angles.
First, we test whether adding multi-hop data to RLVR training based on the original RLVR data improves general vision-language reasoning across model scales and benchmark families (\cref{sec:exp-main-results}).
Second, we examine whether the full multi-hop formulation is necessary, rather than weaker single-hop or truncated-hop variants (\cref{sec:exp-hop-structure-ablation}).
Third, we analyze whether the observed gains align with our central claims about long-CoT robustness, data difficulty coverage, and broad correction of diverse failure modes (\cref{sec:exp-analysis}).

\subsection{Experimental Setup}
We evaluate two models, \textbf{Qwen3.5-35B-A3B} and \textbf{Qwen3.5-397B-A17B}~\CiteParen{qwen35blog}, under three settings: \texttt{Before RLVR}, \texttt{RLVR w/o Multi-Hop}, and \texttt{RLVR w/ Multi-Hop}.
Here, \texttt{Before RLVR} denotes the model after SFT and before RLVR, which serves as the shared starting point for both RLVR settings; \texttt{RLVR w/o Multi-Hop} denotes RLVR training on the original RLVR data only, while \texttt{RLVR w/ Multi-Hop} denotes RLVR training on a mixture of the original RLVR data and our synthesized multi-hop data.
Unless otherwise stated, the SFT data and the original RLVR data used in these experiments are pre-final internal versions, rather than the final official Qwen3.5 data recipe.
For Qwen3.5-35B-A3B, each rollout samples 16 responses for each of 256 queries, uses a mini-batch size of 64 queries, and runs for 1000 gradient steps.
Qwen3.5-397B-A17B follows the same rollout configuration in general, except that the mini-batch size is increased to 128 queries and the total number of gradient steps is 800.
For both models, RLVR training uses the SAPO algorithm with a learning rate of $2.0 \times 10^{-6}$, while other hyperparameters follow the original SAPO paper~\CiteParen{sapo}.

\begin{table}[t]
\centering
\caption{Results of Qwen3.5-35B-A3B. \texttt{Before RLVR} denotes the model after SFT but before RLVR. \texttt{RLVR w/o Multi-Hop} denotes RLVR training on the original RLVR data only. \texttt{RLVR w/ Multi-Hop} denotes RLVR training on the original RLVR data plus the multi-hop data synthesized by HopChain.}
\vspace{-5pt}
\label{tab:exp-35b}
{\renewcommand{\arraystretch}{1.10}
\begin{tabularx}{\linewidth}{@{}>{\raggedright\arraybackslash}p{0.18\linewidth}YYY@{}}
\toprule
\ExperimentHeaderRow
\midrule
\rowcolor{black!3}
\multicolumn{4}{l}{\textbf{STEM and Puzzle}} \\
MathVision & 61.97 & 73.71 & \best{76.05} \\
MMMU Pro & 66.07 & 69.25 & \best{70.64} \\
MMMU & 75.89 & \best{78.89} & 78.33 \\
Mathvista(mini) & 82.80 & \best{85.50} & 85.00 \\
BabyVision & 14.95 & 21.91 & \best{22.68} \\
ZEROBench & 1 & 1 & \best{3} \\
EMMA(mini) & 41.88 & 53.00 & \best{58.00} \\
LogicVista & 63.85 & 74.66 & \best{75.56} \\
\midrule
\rowcolor{black!3}
\multicolumn{4}{l}{\textbf{General VQA}} \\
MMBench$_{\scriptscriptstyle\text{CN-DEV-V1.1}}$ & 87.46 & 90.17 & \best{90.48} \\
MMBench$_{\scriptscriptstyle\text{EN-DEV-V1.1}}$ & 88.47 & 90.63 & \best{91.49} \\
RealWorldQA & 75.42 & 78.17 & \best{79.35} \\
MMStar & 75.20 & 78.53 & \best{78.60} \\
HallusionBench & 66.49 & \best{66.64} & 66.50 \\
AI2D\_TEST & 88.44 & 90.87 & \best{91.29} \\
ERQA & 44.50 & 48.25 & \best{51.38} \\
\midrule
\rowcolor{black!3}
\multicolumn{4}{l}{\textbf{Text Recognition and Document Understanding}} \\
CharXiv & 61.30 & 69.00 & \best{73.10} \\
DocVQA\_VAL & 95.00 & 95.13 & \best{95.55} \\
InfoVQA\_VAL & 86.81 & 87.44 & \best{90.17} \\
\midrule
\rowcolor{black!3}
\multicolumn{4}{l}{\textbf{Video Understanding}} \\
VideoMME$_{\text{w/o sub.}}$ & 73.41 & 74.63 & \best{75.00} \\
VideoMMMU & 70.67 & 73.33 & \best{74.78} \\
MMVUCOT & 63.70 & 65.80 & \best{68.90} \\
MVBench & 69.18 & 69.95 & \best{70.73} \\
LVBench & 51.13 & \best{54.49} & 53.20 \\
MLVU (M-Avg) & 77.92 & 77.69 & \best{79.53} \\
\bottomrule
\end{tabularx}}
\end{table}

\begin{table}[t]
\centering
\caption{Results of Qwen3.5-397B-A17B, the larger model counterpart of \cref{tab:exp-35b}. \texttt{Before RLVR} denotes the model after SFT but before RLVR. \texttt{RLVR w/o Multi-Hop} denotes RLVR training on the original RLVR data only. \texttt{RLVR w/ Multi-Hop} denotes RLVR training on the original RLVR data plus the multi-hop data synthesized by HopChain.}
\vspace{-5pt}
\label{tab:exp-397b}
{\renewcommand{\arraystretch}{1.10}
\begin{tabularx}{\linewidth}{@{}>{\raggedright\arraybackslash}p{0.18\linewidth}YYY@{}}
\toprule
\ExperimentHeaderRow
\midrule
\rowcolor{black!3}
\multicolumn{4}{l}{\textbf{STEM and Puzzle}} \\
MathVision & 77.38 & 81.68 & \best{83.71} \\
MMMU Pro & 73.03 & 75.06 & \best{76.47} \\
MMMU & 79.78 & 81.67 & \best{82.89} \\
Mathvista(mini) & 87.50 & 88.30 & \best{89.00} \\
BabyVision & 17.01 & 28.61 & \best{32.22} \\
ZEROBench & 3 & 4 & \best{8} \\
EMMA(mini) & 58.13 & 66.25 & \best{69.00} \\
LogicVista & 75.62 & 80.69 & \best{81.59} \\
\midrule
\rowcolor{black!3}
\multicolumn{4}{l}{\textbf{General VQA}} \\
MMBench$_{\scriptscriptstyle\text{CN-DEV-V1.1}}$ & 89.47 & 91.41 & \best{91.72} \\
MMBench$_{\scriptscriptstyle\text{EN-DEV-V1.1}}$ & 90.71 & \best{92.49} & 91.56 \\
RealWorldQA & 79.87 & 79.87 & \best{81.70} \\
MMStar & 80.00 & \best{81.73} & 80.67 \\
HallusionBench & 65.76 & 67.48 & \best{67.86} \\
AI2D\_TEST & 91.45 & 92.81 & \best{92.97} \\
ERQA & 53.75 & \best{60.50} & 60.00 \\
\midrule
\rowcolor{black!3}
\multicolumn{4}{l}{\textbf{Text Recognition and Document Understanding}} \\
CharXiv & 70.00 & 74.60 & \best{77.20} \\
DocVQA\_VAL & 95.93 & 95.98 & \best{96.03} \\
InfoVQA\_VAL & 90.42 & 90.83 & \best{92.20} \\
\midrule
\rowcolor{black!3}
\multicolumn{4}{l}{\textbf{Video Understanding}} \\
VideoMME$_{\text{w/o sub.}}$ & 76.56 & 78.30 & \best{80.41} \\
VideoMMMU & 76.67 & 78.89 & \best{80.00} \\
MMVUCOT & 70.20 & 72.30 & \best{72.50} \\
MVBench & 69.63 & 73.03 & \best{73.31} \\
LVBench & 58.30 & \best{59.13} & 59.07 \\
MLVU (M-Avg) & 81.46 & 82.43 & \best{82.52} \\
\bottomrule
\end{tabularx}}
\end{table}

\paragraph{Image Filtering.}
Before multi-hop query synthesis, we filter images to retain cases that are genuinely useful for long vision-language reasoning.
The selection prompt in Appendix~\ref{app:image-select-prompt} is designed to favor images that are perceptually challenging for standard vision models, such as those involving occlusion, dense but still analyzable objects, unusual poses, complex interactions, fine-grained distinctions, or difficult lighting, while filtering out images that are too low-quality or too annotation-impractical to support verifiable reasoning.
To balance quality and throughput, we use a two-stage pipeline.
We first apply Qwen3-VL-235B-A22B-Thinking to a small subset of images with the image-selection prompt and keep the accepted images as the initial set of selected images.
We then use this initial set of selected images together with the large model's outputs to perform supervised fine-tuning (SFT) on a smaller model, Qwen3-VL-30B-A3B-Thinking, specialized for image filtering.
Next, we run this smaller model over all remaining images to obtain a coarse screening result.
Finally, we send the images retained by the coarse screening back to Qwen3-VL-235B-A22B-Thinking for a second, finer filtering pass, yielding a second set of selected images.
The union of the initial set of selected images and the second set of selected images forms the final pool of images used for multi-hop data construction.
Based on the filtered image pool above, we construct multi-hop queries with HopChain and then filter out overly easy samples using a weaker model and the pre-RLVR model, yielding about $6k$--$8k$ multi-hop RLVR samples for each model.
In addition, we use a similar amount of math RLVR data as the original RLVR data.

\paragraph{Benchmarks.}
We evaluate 24 benchmarks across 4 categories.
For \textbf{STEM and puzzle} reasoning, we use MathVision~\CiteParen{wang2024mathvision}, MMMU-Pro~\CiteParen{yue2025mmmupro}, MMMU~\CiteParen{yue2024mmmu}, MathVista~\CiteParen{lu2024mathvista}, BabyVision~\CiteParen{chen2026babyvision}, ZeroBench~\CiteParen{roberts2025zerobench}, EMMA~\CiteParen{hao2025emma}, and LogicVista~\CiteParen{xiao2024logicvista}.
For \textbf{general VQA}, we use MMBench~\CiteParen{liu2023mmbench}, RealWorldQA~\CiteParen{xai2024realworldqa}, MMStar~\CiteParen{chen2024mmstar}, HallusionBench~\CiteParen{guan2024hallusionbench}, AI2D~\CiteParen{kembhavi2016diagram}, and ERQA~\CiteParen{deepmind2025erqa}.
For \textbf{text recognition and document understanding}, we use CharXiv~\CiteParen{wang2024charxiv}, DocVQA~\CiteParen{mathew2021docvqa}, and InfographicVQA (reported as \texttt{InfoVQA\_VAL})~\CiteParen{mathew2021infographicvqa}.
For \textbf{video understanding}, we use Video-MME~\CiteParen{fu2025videomme}, Video-MMMU~\CiteParen{hu2025videommmu}, MVBench~\CiteParen{li2024mvbench}, LVBench~\CiteParen{wang2024lvbench}, MLVU~\CiteParen{zhou2024mlvu}, and MMVU (reported as \texttt{MMVUCOT})~\CiteParen{zhao2025mmvu}.

\subsection{Main Benchmark Results}
\label{sec:exp-main-results}
Tables~\ref{tab:exp-35b} and~\ref{tab:exp-397b} report the main benchmark results on Qwen3.5-35B-A3B and Qwen3.5-397B-A17B, respectively.
RLVR on the original RLVR data (i.e., \texttt{RLVR w/o Multi-Hop}) already improves the \texttt{Before RLVR} models on many benchmarks, but augmenting the original RLVR data with the multi-hop data synthesized by HopChain (i.e., \texttt{RLVR w/ Multi-Hop}) yields further gains on the majority of benchmarks for both model scales.
These gains are broad rather than isolated.
Importantly, our multi-hop data is not synthesized to target any specific benchmark. Even so, it delivers broad gains on the majority of benchmarks, suggesting that HopChain improves generalizable vision-language reasoning rather than overfitting to benchmark-specific patterns.
Quantitatively, \texttt{RLVR w/ Multi-Hop} improves 20 out of 24 benchmarks for both Qwen3.5-35B-A3B and Qwen3.5-397B-A17B.
For Qwen3.5-35B-A3B, the gains cover 6/8 STEM-and-puzzle benchmarks (MathVision, MMMU-Pro, BabyVision, ZeroBench, EMMA, and LogicVista), 6/7 general-VQA benchmarks (MMBench-CN, MMBench-EN, RealWorldQA, MMStar, AI2D, and ERQA), all 3 text-and-document benchmarks (CharXiv, DocVQA, and InfoVQA), and 5/6 video benchmarks (Video-MME, VideoMMMU, MMVUCOT, MVBench, and MLVU).
For Qwen3.5-397B-A17B, the gains are similarly broad: all 8 STEM-and-puzzle benchmarks improve (MathVision, MMMU-Pro, MMMU, MathVista, BabyVision, ZeroBench, EMMA, and LogicVista), along with 4/7 general-VQA benchmarks (MMBench-CN, RealWorldQA, HallusionBench, and AI2D), all 3 text-and-document benchmarks (CharXiv, DocVQA, and InfoVQA), and 5/6 video benchmarks (Video-MME, VideoMMMU, MMVUCOT, MVBench, and MLVU).
These broad quantitative gains are also reflected qualitatively: \Cref{fig:qualitative_examples} shows representative cases in which adding multi-hop data corrects failures made by \texttt{RLVR w/o Multi-Hop}.
Notably, although our multi-hop data is synthesized from images, the improvement transfers strongly to video understanding: video benchmarks improve on 5 out of 6 tasks for both model scales, indicating substantial cross-domain generalization rather than narrow overfitting to image-only training signals.
Across the two tables, the improvements cover all four benchmark families, with only a small number of regressions, indicating that multi-hop data acts as a composable RLVR training signal rather than a narrow task-specific trick.

\subsection{Ablation on Hop Structure: Single-Hop vs. Half-Multi-Hop vs. Multi-Hop}
\label{sec:exp-hop-structure-ablation}
To test whether preserving complex, long hop chains during training is necessary, we compare three training-query settings on the subset of benchmarks shared by all three runs.
In \texttt{RLVR w/ Single Hop}, each multi-hop training query is reduced to only its final hop; in \texttt{RLVR w/ Half-Multi-Hop}, the first half of the chain is removed and only the latter half is kept; and in \texttt{RLVR w/ Multi-Hop}, the full query is kept.

\newcommand{\MethodSingleHop}{\shortstack[c]{RLVR w/\\Single Hop}}
\newcommand{\MethodHalfHop}{\shortstack[c]{RLVR w/\\Half-Multi-Hop}}
\newcommand{\MethodMultiHop}{\shortstack[c]{RLVR w/\\Multi-Hop}}
\newlength{\HopBenchmarkColWidth}
\newlength{\HopResultColWidth}
\setlength{\HopBenchmarkColWidth}{0.28\linewidth}
\setlength{\HopResultColWidth}{\dimexpr(\linewidth-\HopBenchmarkColWidth)/3\relax}
\newcommand{\HopComparisonHeaderRow}{%
\multicolumn{4}{@{}l@{}}{%
  \begingroup
  \setlength{\fboxsep}{0pt}%
  \colorbox{black!5}{%
    \parbox{\linewidth}{%
      \strut
      \makebox[\HopBenchmarkColWidth][l]{\parbox[c][2.6\baselineskip][c]{\HopBenchmarkColWidth}{\raggedright Benchmark}}%
      \makebox[\HopResultColWidth][c]{\parbox[c][2.6\baselineskip][c]{\HopResultColWidth}{\centering \MethodSingleHop}}%
      \makebox[\HopResultColWidth][c]{\parbox[c][2.6\baselineskip][c]{\HopResultColWidth}{\centering \MethodHalfHop}}%
      \makebox[\HopResultColWidth][c]{\parbox[c][2.6\baselineskip][c]{\HopResultColWidth}{\centering \MethodMultiHop}}%
      \strut
    }%
  }%
  \endgroup
}%
\\}

\begin{figure}[t]
\centering
\includegraphics[width=\textwidth]{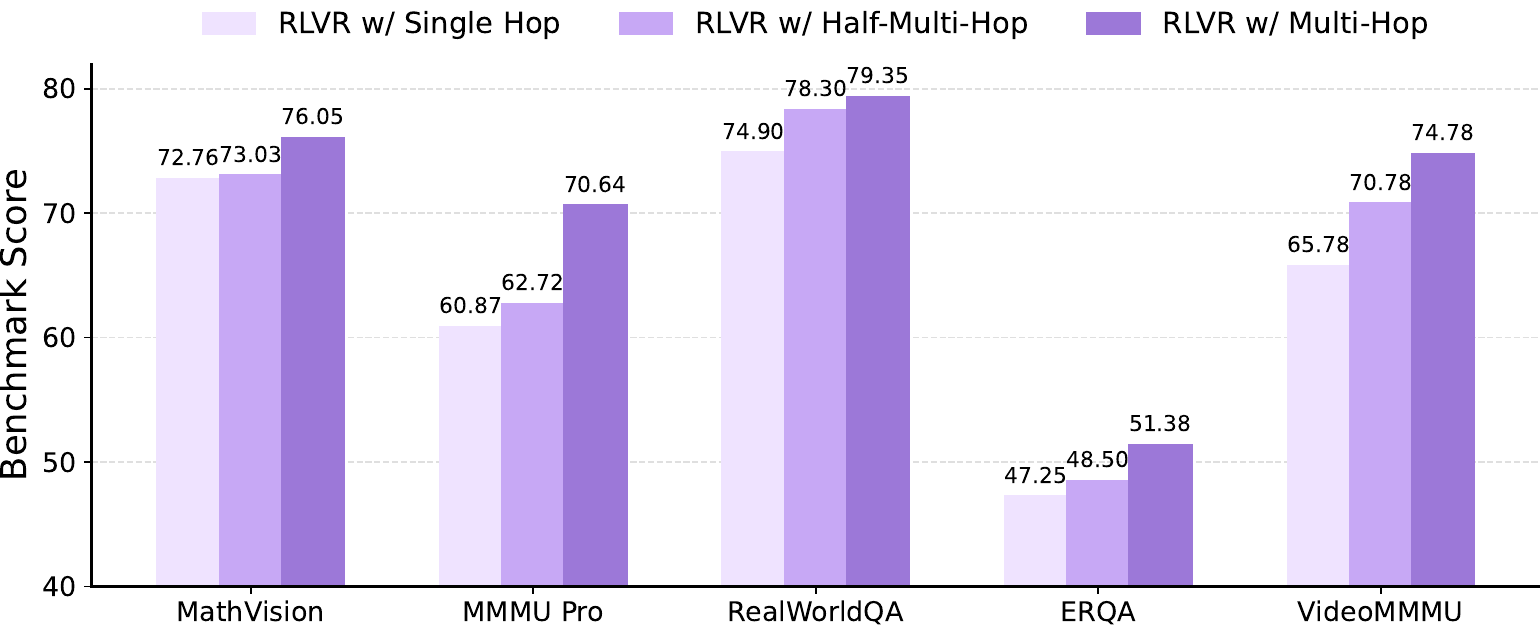}
\caption{Benchmark-level comparison on Qwen3.5-35B-A3B under three training-query settings. \texttt{RLVR w/ Single Hop} simplifies each multi-hop training query to only its final hop, \texttt{RLVR w/ Half-Multi-Hop} removes the first half of the hops and keeps only the latter half, and \texttt{RLVR w/ Multi-Hop} uses the full multi-hop training queries. We evaluate the resulting models on five representative benchmarks, namely MathVision, MMMU Pro, RealWorldQA, ERQA, and VideoMMMU, and plot the benchmark score for each setting. Across all five benchmarks, the full multi-hop setting performs best, showing that preserving the complete multi-hop structure during training is more effective than shortening the query chain.}
\label{fig:table3-representive-benchmarks}
\end{figure}

A benchmark-level comparison on five representative tasks is shown in \cref{fig:table3-representive-benchmarks}.
On MathVision, MMMU Pro, RealWorldQA, ERQA, and VideoMMMU, the ordering is consistent: \texttt{RLVR w/ Multi-Hop} performs best, followed by \texttt{RLVR w/ Half-Multi-Hop}, and then \texttt{RLVR w/ Single Hop}.
This per-benchmark ordering is also reflected in the aggregate: averaged across these five representative benchmarks, the score drops from 70.4 for \texttt{RLVR w/ Multi-Hop} to 66.7 for \texttt{RLVR w/ Half-Multi-Hop} and 64.3 for \texttt{RLVR w/ Single Hop}.
This pattern shows that the benefit does not come from collapsing the training queries to the last hop or to a shortened chain; preserving longer cross-hop dependencies is important.

\subsection{Analysis}
\label{sec:exp-analysis}
\begin{figure}[t]
\centering
\includegraphics[width=\textwidth]{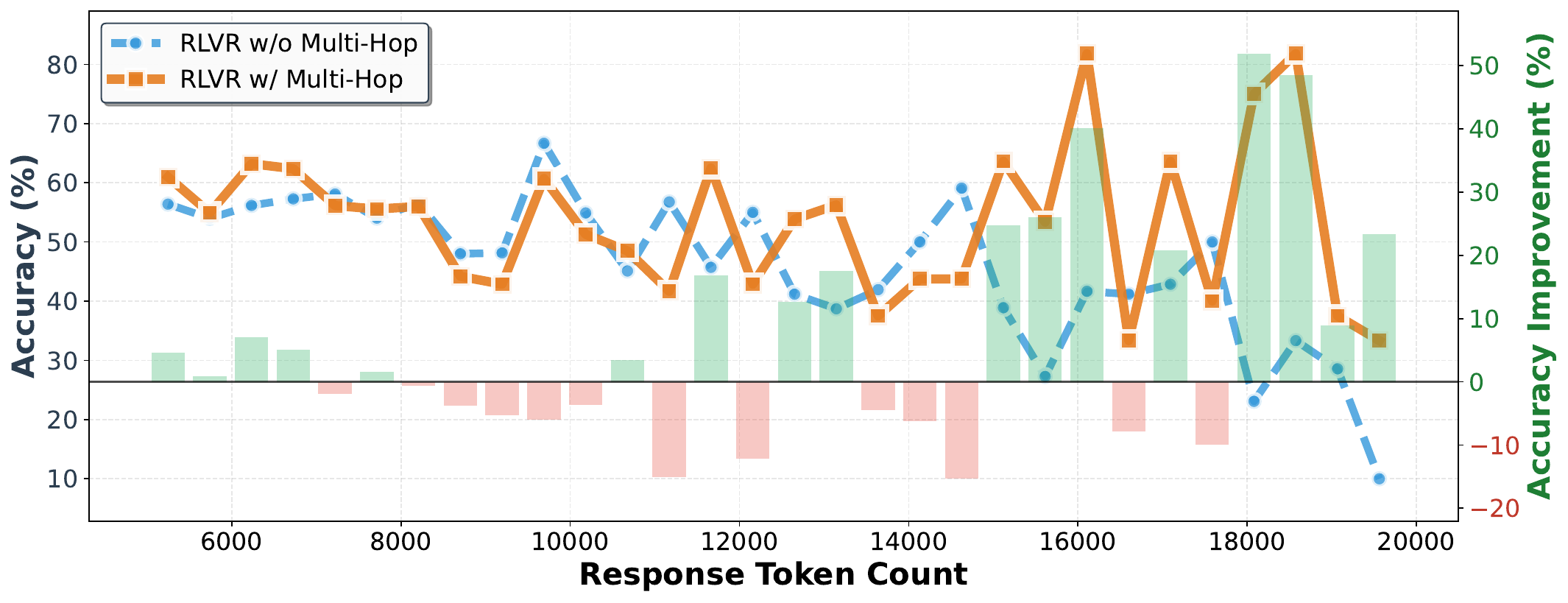}
\caption{Performance as a function of response length, aggregated from the benchmark evaluation results of Qwen3.5-397B-A17B in \cref{tab:exp-397b}. The blue dashed line and the orange solid line show the accuracies of \texttt{RLVR w/o Multi-Hop} and \texttt{RLVR w/ Multi-Hop}, respectively, across bins of increasing response token count, while the bars show the corresponding accuracy improvement of \texttt{RLVR w/ Multi-Hop} over \texttt{RLVR w/o Multi-Hop} on the right axis. The advantage of multi-hop training remains evident on long responses and is especially pronounced in the ultra-long-CoT region.}
\label{fig:ultra_long_cot_advantage}
\end{figure}

\paragraph{Analysis by Reasoning Length.}
If HopChain primarily strengthens long-CoT reasoning, its advantage should persist as response chains become longer.
\Cref{fig:ultra_long_cot_advantage} tests exactly this on Qwen3.5-397B-A17B by binning benchmark responses by response token count.
The advantage of \texttt{RLVR w/ Multi-Hop} over \texttt{RLVR w/o Multi-Hop} remains visible across the bins and is often larger in the ultra-long-response regime.
This supports the interpretation that HopChain improves robust chained reasoning over long outputs, rather than only helping on short or easy responses.

\begin{figure}[t]
\centering
\includegraphics[width=\textwidth]{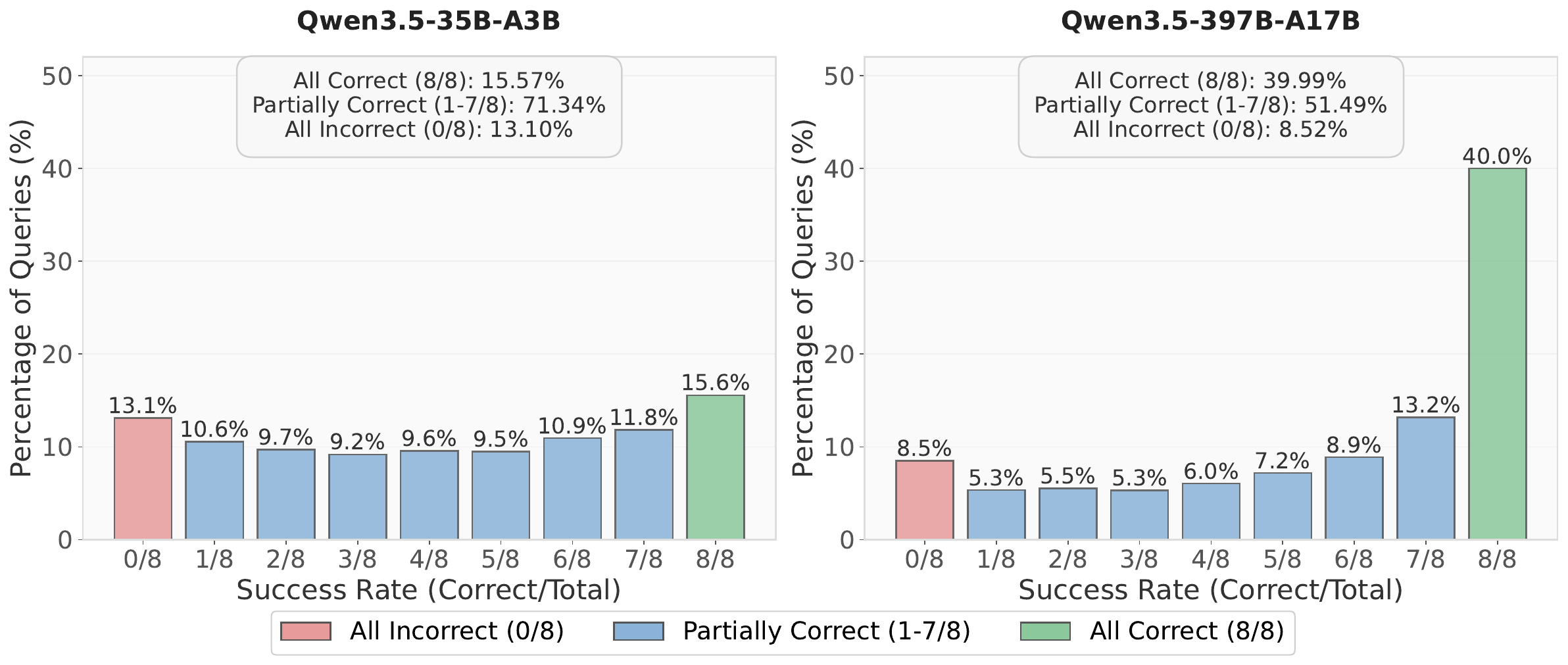}
\caption{Query-level success-rate distributions for Qwen3.5-35B-A3B and Qwen3.5-397B-A17B. For each multi-hop query, we independently sample eight responses and bucket the query by how many of the eight are correct, from \texttt{0/8} to \texttt{8/8}. For both models, more than half of the queries fall into the \texttt{Partially Correct} regime, and the queries are distributed across multiple success buckets. This suggests that the synthesized multi-hop data spans a broad difficulty range and is suitable for RLVR training across models of different sizes and capability levels.}
\label{fig:turbo_vs_plus_success_rate_histogram}
\end{figure}

\paragraph{Difficulty Coverage Across Model Scales.}
A useful multi-hop training set should cover a broad range of difficulties, rather than collapsing to queries that are either trivial or essentially unsolvable.
\Cref{fig:turbo_vs_plus_success_rate_histogram} measures this by independently sampling eight responses for each multi-hop query and grouping the query by how many sampled responses are correct.
For both Qwen3.5-35B-A3B and Qwen3.5-397B-A17B, more than half of the queries fall into the \texttt{Partially Correct} regime, and the distribution spans multiple success buckets.
This indicates that the synthesized multi-hop data covers a broad difficulty range and can provide a useful RLVR training signal to models of different sizes and capability levels.

\begin{figure}[t]
\centering
\includegraphics[width=\textwidth]{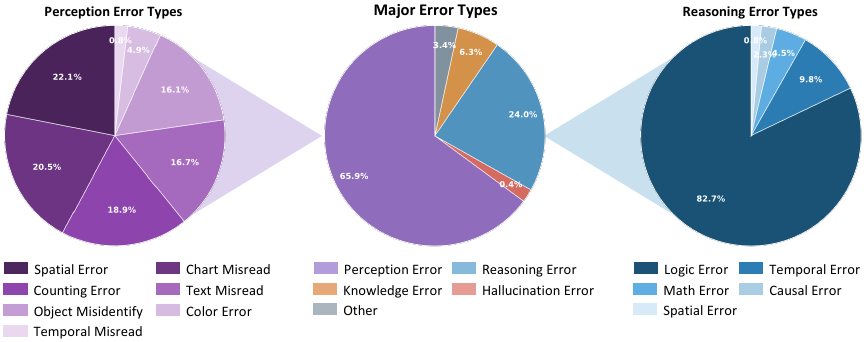}
\caption{Error distribution of the original \texttt{RLVR w/o Multi-Hop} errors among the cases where \texttt{RLVR w/ Multi-Hop} improves over \texttt{RLVR w/o Multi-Hop} on the benchmarks in \cref{tab:exp-35b,tab:exp-397b}. \textbf{Center pie chart:} the major error categories of the corrected cases, indicating which types of baseline errors are most often fixed after adding multi-hop data. \textbf{Left pie chart:} the subtype breakdown of corrected perception errors. \textbf{Right pie chart:} the subtype breakdown of corrected reasoning errors. With \cref{fig:error_distribution}, this figure shows that the multi-hop data synthesized by HopChain can improve models in broad, generalizable domains.}
\label{fig:improve_error_distribution}
\end{figure}

\paragraph{Error-Type Analysis.}
Finally, we ask whether multi-hop augmentation fixes only a narrow corner of the error space or yields broader gains.
\Cref{fig:error_distribution} analyzes the original error types of the cases where \texttt{RLVR w/ Multi-Hop} corrects an error made by \texttt{RLVR w/o Multi-Hop}.
Compared with the baseline error profile in \cref{fig:error_distribution}, the corrected-case distribution is broadly similar: perception errors remain the largest group, reasoning errors remain the second largest, and knowledge, hallucination, and other errors are also represented.
The subtype breakdowns in \Cref{fig:improve_error_distribution} further show that these improvements extend across diverse subcategories, covering chart and text misreads, object misidentification, spatial and counting errors, as well as logic, math, temporal, and causal reasoning errors.
Taken together, the close correspondence to the baseline error distribution in \cref{fig:error_distribution} and the broad subtype gains in \cref{fig:improve_error_distribution} suggest that HopChain does not only patch a single narrow failure type; instead, it yields broad gains across the diverse failure modes in long-CoT vision-language reasoning.

\section{Related Work}

\subsection{Vision-Language Models}

The integration of visual encoders with large language models has driven rapid progress in multimodal understanding.
LLaVA~\CiteParen{liu2024visual} introduced the visual instruction tuning paradigm by projecting visual features into the language model's embedding space, a design subsequently refined in LLaVA-1.5~\CiteParen{liu2024improved} with stronger projectors and higher-resolution inputs.
This architectural template has been adopted and scaled by both open-source and proprietary efforts, including InternVL~\CiteParen{chen2024internvl,chen2024internvl2}, Qwen-VL~\CiteParen{bai2023qwenvl,bai2025qwen25vl}, GPT-4V~\CiteParen{openai2023gpt4}, and Gemini~\CiteParen{geminiteam2024gemini}.
Alongside these advances, diagnostic studies show that even strong VLMs remain vulnerable to object hallucination, visual illusion, and language-prior-driven errors~\CiteParen{rohrbach2018object,guan2024hallusionbench,leng2024vcd}.
Recent analyses further suggest that these failures can become more pronounced when multimodal reasoning chains grow longer and drift away from image-grounded evidence~\CiteParen{liu2025morethinking}.
While these models achieve strong aggregate scores on diverse benchmarks, their visual perception can still become unreliable during multi-step reasoning.
This is the bottleneck our work targets through synthesized multi-hop data.

\subsection{Reinforcement Learning for Language and Vision-Language Models}

Reinforcement learning from human feedback (RLHF)~\CiteParen{ouyang2022training} first demonstrated the effectiveness of reinforcement learning (RL)-based alignment for language models, using proximal policy optimization (PPO)~\CiteParen{schulman2017proximal} as the default optimizer.
More recently, reinforcement learning with verifiable rewards (RLVR)~\CiteParen{shao2024deepseekmath} has emerged as a powerful alternative to RLHF, eliminating the need for a learned reward model by using programmatically verifiable answers.
DeepSeek-R1~\CiteParen{deepseekr1} further showed that pure RL can induce strong chain-of-thought reasoning in LLMs, which has motivated parallel extensions to VLMs such as VLM-R1~\CiteParen{shen2025vlmr1} and TikArt~\CiteParen{ding2026tikart}.
In parallel, the optimization toolkit has also evolved: GRPO~\CiteParen{shao2024deepseekmath} and GSPO~\CiteParen{gspo} use group-based advantage estimation with hard clipping, while SAPO~\CiteParen{sapo} replaces hard clipping with a temperature-controlled soft gate for improved stability.
Beyond algorithm design, recent mechanistic and robustness analyses suggest that RL's effects are selective rather than uniform. In LLMs, improvements are concentrated on a minority of high-entropy tokens~\CiteParen{wang2025beyond}; in VLMs, RL appears to primarily refine vision-to-reasoning alignment in mid-to-late layers rather than uniformly strengthening visual perception~\CiteParen{li2026frankenstein}, and RL-finetuned models can still exhibit weak visual grounding and over-reliance on textual cues~\CiteParen{zhao2026robustness}.
In related multimodal reasoning settings, recent work has also begun to explicitly reward visually grounded reasoning traces, for example through visually grounded reinforcement finetuning for visual document reasoning~\CiteParen{ni2025pointrft} and evidence-grounded RL for video reasoning~\CiteParen{luo2025thinkingdrifts}.
Most existing RL and RLVR works remain tied to pre-existing task or benchmark data, including math- and science-oriented reasoning sets as well as task-specific multimodal datasets for documents or videos.
In contrast, our pipeline generates a benchmark-agnostic proxy task designed to force repeated visual grounding and yield broad, generalizable improvements.

\subsection{Vision-Language Reasoning and Multi-Hop Reasoning}

Compositional visual reasoning has been studied through synthetic diagnostic datasets such as CLEVR~\CiteParen{johnson2017clevr} and real-image benchmarks such as GQA~\CiteParen{hudson2019gqa}, with early architectural innovations including neural module networks~\CiteParen{andreas2016neural} and visual programming~\CiteParen{gupta2023visual}.
In the language domain, multi-hop question answering requires chaining evidence across multiple passages.
This setting has been formalized by benchmarks such as HotpotQA~\CiteParen{yang2018hotpotqa}.
Chain-of-thought (CoT) prompting~\CiteParen{wei2022chain} and its multimodal extension~\CiteParen{zhang2023multimodal} have shown that eliciting step-by-step reasoning significantly improves both LLM and VLM performance.
At the same time, several recent works argue that strong multimodal reasoning requires more than language-space CoT: it depends on fine-grained observation, stronger intermediate perception representations, repeated image inspection, or iterative revisiting of visual regions~\CiteParen{bigverdi2025perception,ye2025blinktwice,jiang2025vlmr3}.
Our work differs from prior visual reasoning research in three respects.
(1)~We formalize multi-hop vision-language reasoning with two complementary hop types: perception-level hops (switching between single-object and multi-object perception) and instance-chain hops (A $\rightarrow$ B $\rightarrow$ C dependency chains), instead of treating it as a single-axis chain.
(2)~We use multi-hop reasoning as a \emph{proxy task} to improve general VLM capabilities rather than as an end goal.
(3)~Our data is synthesized on real images at scale, which bridges the gap between synthetic diagnostics and manually curated benchmarks.

\subsection{Scalable Data Synthesis for Model Training}

Using strong models to generate training data for weaker ones has become a central scaling strategy.
Self-Instruct~\CiteParen{wang2023selfinstruct} and Alpaca~\CiteParen{taori2023stanford} pioneered LLM-generated instruction data, while ShareGPT4V~\CiteParen{chen2023sharegpt4v} extended this idea to vision-language captions.
Representative multimodal data synthesis pipelines additionally integrate vision foundation models.
They often combine a VLM for object detection with SAM/SAM2~\CiteParen{kirillov2023segment,ravi2024sam2} for instance segmentation.
They may also use open-set detection approaches such as Grounding DINO~\CiteParen{liu2023grounding} to construct structured multi-hop queries from raw images.
Unlike prior data synthesis efforts that aim to approximate the target task distribution, our pipeline generates a benchmark-agnostic proxy task designed to force repeated visual grounding for generalizable improvement.

\section{Conclusion}

In this paper, we identify diverse and compounding failure modes during long CoT reasoning as a central barrier to robust vision-language reasoning. To address this challenge, we propose HopChain, a scalable framework for synthesizing multi-hop vision-language reasoning data for RLVR training, where earlier hops establish the instances, sets, or conditions needed for later hops, forcing repeated visual re-grounding throughout training while terminating in specific, unambiguous numerical answers suitable for verifiable rewards. By augmenting the original RLVR data with HopChain-synthesized multi-hop data, we obtain broad and generalizable gains on 20 out of 24 benchmarks on both Qwen3.5-35B-A3B and Qwen3.5-397B-A17B. Importantly, these gains arise even though the synthesized data is benchmark-agnostic rather than designed for any specific benchmark. Additional analyses further show that preserving full chained queries is important, that multi-hop data substantially strengthens long-CoT vision-language reasoning, and that the synthesized data spans a broad difficulty range while enabling trained models to correct a broad range of errors. These results establish HopChain as an effective and scalable framework for improving generalizable vision-language reasoning in VLMs.

Despite these gains, the current pipeline still depends on successful instance segmentation, so images with no detectable objects (and thus no SAM3-segmentable instances) cannot be processed and are excluded from the current synthesis workflow. A natural next step is to reduce this dependency by introducing complementary data-construction routes for images with few or no segmentable objects, while preserving the core design principle of chained visual grounding for long-CoT RLVR training.

\newpage
{
\small
\bibliographystyle{plainnat}
\bibliography{ref.bib}
}

\newpage
\appendix
\renewcommand{\theHsection}{app.\thesection}
\section*{Appendix}

\section{Full Prompt for Multi-Hop Query Design}
\label{app:multihop-prompt}

We provide the complete prompt used for synthesizing multi-hop vision-language reasoning queries below. Placeholders such as \texttt{\{num\_queries\_word\}}, \texttt{\{object\_list\}}, \texttt{\{num\_queries\}}, and \texttt{\{target\_hop\_count\_info\}} are filled at runtime.

\medskip
\noindent
\begin{tcolorbox}[
  breakable,
  title={\textbf{Multi-Hop Query Design Prompt (Full)}},
  fonttitle=\bfseries,
  colback=white,
  colframe=black,
  boxrule=0.5pt,
  left=4pt,
  right=4pt,
  top=6pt,
  bottom=6pt,
]
\lstset{
  basicstyle=\small\ttfamily,
  breaklines=true,
  columns=flexible,
  frame=none,
  numbers=none,
  showstringspaces=false,
  keepspaces=true,
  breakatwhitespace=false,
  xleftmargin=0pt,
  aboveskip=0pt,
  belowskip=0pt,
}
\begin{lstlisting}
#### **Role & Goal**

You are a top-tier AI multimodal capability evaluation expert. Your task is to design a set of **{num_queries_word}** independent, high-difficulty **multi-hop vision-language reasoning** sub-queries based on a **complex image**. These sub-queries will serve as "capability modules" to evaluate a model's comprehensive visual intelligence.

#### **Understanding VLM Perception Capabilities**

VLM (Vision-Language Model) perception capabilities can be categorized into three levels:

1. **Single-Object Perception (Level 1):** Perceiving attributes of a single object (e.g., color, shape, size, text content, position, category).
2. **Multi-Object & Relationship Perception (Level 2):** Perceiving multiple objects and their relationships (e.g., spatial relationships like "A is to the left of B", comparative relationships like "A is larger than B", counting objects that satisfy certain conditions).
3. **Multi-Hop Reasoning (Level 3):** Chaining multiple perception steps together. **Multi-hop can occur in TWO dimensions:**
   - **Perception-Level Hops:** Moving between Level 1 and Level 2 perception tasks
   - **Instance-to-Instance Hops:** Reasoning from instance_1 -> instance_2 -> instance_3 -> ... -> instance_n, where **finding the next instance DEPENDS ON the previous instance**. The key is that instance_N+1 can ONLY be located by using information from or relationships with instance_N.

**YOUR TASK: Design queries that are strictly Level 3 (Multi-Hop Reasoning).** Each query must chain together multiple perception steps across BOTH dimensions - hopping between different perception levels AND hopping across different instances with DEPENDENCY relationships.

#### **Types of Multi-Hop Reasoning (Design queries using ALL types)**

**Type A - Instance Dependency Chain (MOST IMPORTANT):** 
- The NEXT instance can ONLY be found by using its relationship with the PREVIOUS instance
- Each hop must establish a dependency: "the X that is [relationship] to the previous instance"
- Example: "Start from the red shirt -> find the chair that THIS shirt is draped over -> find the table that THIS chair is placed next to -> count objects on THIS table"
- Example: "Find person A -> locate the bag that A is holding -> identify the brand logo on THIS bag -> count similar logos in the image"
- **BAD (no dependency):** "Find the largest car, then find the tallest tree" (tree doesn't depend on car)
- **GOOD (with dependency):** "Find the largest car, then find the tree closest to THIS car" (tree depends on which car was found)

**Type B - Perception Level Hop:**
- Level 1 (single object) -> Level 2 (multi-object relationship) -> Level 1 -> Level 2 -> ...
- Each level transition should maintain the instance dependency chain
- Example: "Read the number on sign A (L1), find objects to the right of THIS sign (L2), among THESE objects find the closest to B (L2), read text on THIS object (L1)"

**Type C - Combined with Strong Dependencies (PREFERRED):**
- Combine instance dependency chains with perception level changes for maximum complexity
- Every instance transition must have a clear dependency relationship
- Example: "Find shirt A -> locate the chair THIS shirt is draped on -> find the person sitting on THIS chair -> read the text on the book THIS person is holding -> count pages visible in THIS book"

**You are provided with MULTIPLE images:**
1. **Image 1 (Original):** The original image containing all visual information. **This is the ONLY image that will be available when answering the queries.**
2. **Instance Patch Images (Image 2, 3, 4, ...):** Cropped patches of each individual instance from the original image. Each patch shows one specific instance extracted using its bounding box. These patches help you precisely identify each instance's appearance and content. **These patches are ONLY for your reference during question design and will NOT be available when answering queries.**

**CRITICAL: You are designing queries for a SPECIFIC COMBINATION of objects. The following object instances are the ONLY instances you should consider for this task. Each instance is provided as a separate cropped patch with its bounding box location in the original image (coordinates in 0-1000 range):**

{object_list}

**[WARNING] MANDATORY REQUIREMENT - INVOLVE ALL INSTANCES:**
- **Each sub-query MUST involve AS MANY instances as possible from the combination list above.**
- **Ideally, every query should reference or require reasoning about ALL or MOST of the listed instances.**
- If the combination has N instances, each query should involve at least (N-1) instances, preferably all N instances.
- The `involved_objects` field in your output MUST list ALL instances that are referenced or reasoned about in that query.
- **DO NOT design queries that only involve 2-3 objects when 5-6 objects are available in the combination.**

**CRITICAL CONSTRAINT - NO PATCH/BOX REFERENCES & UNAMBIGUOUS INSTANCE REFERENCES:**
- The cropped patches and bounding box coordinates are **ONLY** visual aids to help you understand which object is which during question design.
- **When writing the actual query text, you MUST NEVER mention:**
  - Detection borders, bounding boxes, patches, or any detection-related visual markers
  - Bounding box coordinates or positions in the 0-1000 coordinate system
  - Any reference to the cropped patch images or instance visualization aids
- **You MUST describe objects using ONLY:**
  - Spatial positions (e.g., "the object on the left", "the topmost item")
  - Contextual relationships (e.g., "the object next to the table", "the item above the text")
  - Functional descriptions (e.g., "the button", "the icon", "the label")
  - Visual attributes from the original image itself (e.g., actual colors of objects, sizes, shapes)
- **[WARNING] CRITICAL: UNAMBIGUOUS INSTANCE REFERENCES (MOST IMPORTANT):**
  - **Each instance reference in your query MUST uniquely identify exactly ONE instance in the original image (Image 1) without any borders.**
  - **Test each reference:** Imagine the query is given to someone looking ONLY at Image 1 (no colored borders). Can they identify the EXACT instance you mean without any confusion?
  - **BAD (ambiguous):** "the car" when there are multiple cars in the image
  - **BAD (ambiguous):** "the person on the right" when there are two people on the right side
  - **BAD (ambiguous):** "the red object" when multiple objects are red
  - **GOOD (unambiguous):** "the car closest to the bottom-left corner"
  - **GOOD (unambiguous):** "the person wearing a blue shirt standing next to the lamppost"
  - **GOOD (unambiguous):** "the largest red object that is positioned above the text area"
  - **GOOD (unambiguous):** "the bicycle located between the tree and the bench"
  - **Use multiple discriminating features:** Combine spatial position + visual attributes + contextual relationships to ensure uniqueness
  - **Verify uniqueness:** For each instance reference, mentally check if ANY other object in Image 1 could match that description. If yes, add more distinguishing details.
- **Reason:** The final questions will be evaluated on Image 1 (the original image) **without any borders, markers, or detection aids**. Any ambiguous reference will make the question unsolvable or lead to multiple valid interpretations.

#### **Core Principles: Requirements for Each Sub-query**

**Every** sub-query you design must strictly adhere to all of the following standards:

1.  **Independence:** Each query must be completely self-contained. Solving it must not depend on information or answers from any other query.

2.  **[VERY IMPORTANT] MANDATORY Multi-Hop Structure (MOST CRITICAL):** 
    - **Each query MUST be a genuine multi-hop reasoning problem with {target_hop_count_info}.**
    - **Number of hops = {target_hop_count_info}.** Push the boundaries of complexity within this range.
    - Each hop can be:
      - **Level 1 (Single-Object Perception):** Extract information from one specific object (e.g., read text, identify color, determine position)
      - **Level 2 (Multi-Object Relationship):** Reason about relationships between multiple objects (e.g., compare sizes, count objects satisfying a condition, determine spatial arrangement)
    - **Instance-to-Instance Chaining is REQUIRED:** The reasoning should hop from instance_1 -> instance_2 -> instance_3 -> ... involving as many instances as possible in a chain.
    - **The hops must be logically chained:** The result of each hop must be REQUIRED for the next hop. No hop should be skippable.
    - **Example of a HIGH-QUALITY multi-hop structure (5 hops, all instances involved):**
      - Hop 1 (L1 on A): Read the number on object A -> get value X
      - Hop 2 (L2 on A,B,C): Find objects to the right of A that are smaller than B -> get set S containing C
      - Hop 3 (L1 on C): Identify the color of object C (from set S) -> get color Y
      - Hop 4 (L2 on D,E,F): Count objects among D,E,F that share color Y -> get count Z
      - Hop 5 (L2 on result): Compute X x Z and compare with the area ratio of E to F -> final answer
    - **INVALID (too few hops):** "How many red objects are there?" (1 hop)
    - **INVALID (too simple):** "What is the color of the largest object?" (2 hops at most)
    - **INVALID (no instance chain):** Query that only touches 2 instances

3.  **[IMPORTANT] Maximum Instance Coverage (CRITICAL):**
    - **Each query MUST involve ALL instances from the provided combination list - no exceptions.**
    - Design queries where EVERY instance plays a meaningful role in the reasoning chain.
    - **Good pattern:** "Starting from A, find the instance closest to it (hop to B), then find instances sharing B's color (hop to C,D), compare their sizes with E (hop to E), and count those larger than F (hop to F) -> final answer uses info from ALL instances"
    - **Bad pattern:** "Compare object 1 and object 2" (ignores other instances)

4.  **[IMPORTANT] High Perception & Logic Difficulty (CRITICAL):**
    - **Perception difficulty:** Each hop should require non-trivial perception:
      - Reading small or partially occluded text
      - Distinguishing subtle color differences
      - Determining complex spatial relationships (e.g., "between", "surrounded by", "aligned with")
      - Counting objects under multiple constraints
    - **Logic difficulty:** The reasoning chain should require:
      - Conditional logic (if-then-else based on visual observations). **IMPORTANT: Ensure that the "if" conditions are NOT always true. Aim for a balanced and random distribution of "Yes" and "No" outcomes for these conditions across different queries and hops to avoid bias.**
      - Set operations (intersection, union, difference of object sets)
      - Numerical computations (ratios, differences, products)
      - Comparative reasoning with multiple criteria
    - **Avoid trivial hops** like "identify that this is a car" - instead require "identify the car that is closest to the pedestrian and has a license plate starting with a letter"

5.  **Deterministic Numerical Answer:** The final answer to each query must be a specific, unambiguous **number**.

6.  **Visual Grounding:** All information required to solve the query must be obtainable **only** from the image and must not rely on external knowledge.

7.  **Avoid "Shortcut" Hints:** Strictly prohibit using low-level visual features (e.g., "the red text," "the largest icon") to locate information. Guidance should be provided through the object's **function, context, or logical relationships**.

8.  **Avoid Visual Information Leakage:** The wording of the query **must not contain** any specific text or numbers that the model would need to acquire from the image via visual recognition (e.g., OCR). This ensures the model finds the answer by understanding the image content, not through text matching.

9.  **Unambiguous Phrasing & Deterministic Solution:** 
    - **The query itself must be phrased precisely, allowing for only one logical, unambiguous interpretation within the context of the image.**
    - **The solution process must be completely deterministic with no room for dispute:**
      - Every instance reference must uniquely identify exactly one object (see CRITICAL CONSTRAINT above)
      - Every spatial relationship must have a clear, unambiguous definition (e.g., "to the left of" means the object's center is to the left of the reference object's center)
      - Every comparison criterion must be objective and measurable (e.g., "larger" means larger area, not perceived importance)
      - Every counting task must have clear inclusion/exclusion criteria
      - Every numerical operation must be well-defined
    - **BAD (ambiguous solution):** "Count objects near the tree" - what distance counts as "near"?
    - **GOOD (deterministic):** "Count objects whose center is within a distance of twice the tree's width from the tree's center"
    - **BAD (ambiguous):** "Find the most prominent object" - subjective judgment
    - **GOOD (deterministic):** "Find the object with the largest area"
    - **Verification test:** Two different people with good vision looking at Image 1 should arrive at the EXACT SAME numerical answer through the EXACT SAME reasoning steps.

10. **Diversity of Reasoning:** For a set of sub-queries designed for the same image, avoid using the same reasoning logic or template. Each sub-query should test different reasoning paths, hop combinations, and instance chains. **This includes diversifying the outcomes of conditional steps (mixing "if true" and "if false" paths) so that the correct answer is neither always "Yes" nor always "No".**

11. **Capability Coverage:** The {num_queries_word} queries you design, **as a whole**, should cover the following core capability dimensions:
    *   **Color Comprehension**
    *   **Spatial Reasoning**
    *   **Counting & Aggregation**
    *   **Text & Symbol Recognition**
    *   **Logical & Conditional Inference**

12. **No Detection Border/Color References:** **STRICTLY FORBIDDEN** to mention detection borders, border colors, bounding boxes, or any detection-related visual markers in the query text.
13. **Balanced Conditional Outcomes (CRITICAL):** For queries involving conditional logic (if-then-else), you MUST ensure a balanced and random distribution of "Yes" and "No" outcomes for the conditions. **Do not make the "if" condition always true or always false.** Across all queries and hops, roughly half of the conditional checks should evaluate to "No" (triggering the 'else' or 'otherwise' path). For example, if an object is red, you might design a condition like "if this object is blue, add 5; otherwise, add 2", forcing the reasoning to follow the "otherwise" path. This prevents the model from developing a bias and ensures it actually performs the visual verification.

#### **Workflow & Output Format**

Your final output should be a **single JSON object**. This object contains an array named `sub_queries`, with {num_queries_word} elements, each representing a meticulously designed sub-query. 

**IMPORTANT: The `involved_objects` field MUST list ALL instances from the combination that are referenced in the query. Aim to involve ALL available instances in each query.**

Please strictly follow the JSON structure below:

"`json
{{
  "sub_queries": [
    {{
      "id": 1,
      "primary_capability": "Spatial Reasoning + Counting + Conditional Logic",
      "involved_objects": ["instance_1", "instance_2", "instance_3", "instance_4", "instance_5"],
      "query": "Your complex multi-hop query here. It should naturally chain through multiple instances: Start from A, hop to B based on some relationship, then hop to C, then to D, and finally involve E to compute the answer.",
      "instance_chain": "A -> B -> C -> D -> E (describe how reasoning flows between instances)",
      "reasoning_hops": [
        {{
          "hop_number": 1,
          "hop_type": "Level 1 (Single-Object)",
          "from_instance": "instance_1",
          "to_instance": null,
          "description": "Extract specific information from instance_1",
          "objects_involved": ["instance_1"],
          "output": "value_X extracted from instance_1"
        }},
        {{
          "hop_number": 2,
          "hop_type": "Level 2 (Multi-Object Relationship)",
          "from_instance": "instance_1",
          "to_instance": "instance_2",
          "description": "Use value_X to find/filter among other instances, identifying instance_2",
          "objects_involved": ["instance_1", "instance_2", "instance_3"],
          "output": "instance_2 identified based on relationship with instance_1"
        }},
        {{
          "hop_number": 3,
          "hop_type": "Level 2 (Multi-Object Relationship)",
          "from_instance": "instance_2",
          "to_instance": "instance_4",
          "description": "From instance_2, determine relationship with instance_4",
          "objects_involved": ["instance_2", "instance_4"],
          "output": "relationship_result between instance_2 and instance_4"
        }},
        {{
          "hop_number": 4,
          "hop_type": "Level 1 (Single-Object)",
          "from_instance": "instance_4",
          "to_instance": null,
          "description": "Extract value from instance_4",
          "objects_involved": ["instance_4"],
          "output": "value_Y from instance_4"
        }},
        {{
          "hop_number": 5,
          "hop_type": "Level 2 (Multi-Object Relationship)",
          "from_instance": "instance_4",
          "to_instance": "instance_5",
          "description": "Compare or compute using instance_5 as reference",
          "objects_involved": ["instance_4", "instance_5"],
          "output": "final_numerical_answer"
        }}
      ],
      "hypothetical_answer": "42",
      "design_rationale": "Explain: (1) The instance chain A->B->C->D->E, (2) Why each hop is necessary, (3) What makes the perception/logic difficult"
    }},
    {{
      "id": 2,
      "primary_capability": "Text Recognition + Logical Inference + Conditional",
      "involved_objects": ["ALL instances - list them all"],
      "query": "Another complex multi-hop query with instance chain...",
      "instance_chain": "Describe the instance flow",
      "reasoning_hops": [
        {{"hop_number": 1, "hop_type": "L1", "from_instance": "...", "to_instance": "...", "description": "...", "objects_involved": [...], "output": "..."}},
        {{"hop_number": 2, "hop_type": "L2", "from_instance": "...", "to_instance": "...", "description": "...", "objects_involved": [...], "output": "..."}},
        {{"hop_number": 3, "hop_type": "...", "from_instance": "...", "to_instance": "...", "description": "...", "objects_involved": [...], "output": "..."}},
        {{"hop_number": 4, "hop_type": "...", "from_instance": "...", "to_instance": "...", "description": "...", "objects_involved": [...], "output": "..."}}
      ],
      "hypothetical_answer": "150",
      "design_rationale": "..."
    }},
    {{
      "id": 3,
      "primary_capability": "Counting & Aggregation + Spatial + Set Operations",
      "involved_objects": ["ALL instances"],
      "query": "Complex multi-hop query...",
      "instance_chain": "...",
      "reasoning_hops": ["... {target_hop_count_info} ..."],
      "hypothetical_answer": "24",
      "design_rationale": "..."
    }},
    {{
      "id": 4,
      "primary_capability": "Color + Comparison + Conditional Branching",
      "involved_objects": ["ALL instances"],
      "query": "Complex multi-hop query...",
      "instance_chain": "...",
      "reasoning_hops": ["... {target_hop_count_info} ..."],
      "hypothetical_answer": "3",
      "design_rationale": "..."
    }},
    {{
      "id": {num_queries},
      "primary_capability": "Comprehensive: All Capabilities Combined",
      "involved_objects": ["ALL instances"],
      "query": "Most complex multi-hop query combining all capability types with longest instance chain...",
      "instance_chain": "...",
      "reasoning_hops": ["... {target_hop_count_info} ..."],
      "hypothetical_answer": "1800",
      "design_rationale": "..."
    }}
  ]
}}
"`

#### **Examples of Well-Designed Multi-Hop Queries**

Suppose the combination contains: person_1, car_1, bicycle_1, sign_1, tree_1

---

**EXCELLENT Example (5 hops, instance chain, all 5 objects, high difficulty, UNAMBIGUOUS references):**

**Context:** Assume in the image: sign is in top-right corner, car is red sedan in center, bicycle is yellow near bottom, person wearing blue shirt in left, tree is large with green leaves in right background.

Query (with unambiguous references): "Starting from the traffic sign located in the top-right corner of the image, read the speed limit number displayed on it. Then find which vehicle - either the red car in the center or the yellow bicycle near the bottom edge - has its center point closest to the sign's center. Check whether that vehicle's center is closer to the person wearing the blue shirt on the left side, or to the large tree with green leaves in the right background. If it's closer to the person, count all objects whose centers lie on the straight line connecting the person and the tree; if closer to the tree, count all objects positioned to the left of the tree's leftmost edge. Finally, multiply the speed limit number by this count, then subtract the total number of wheels visible on both vehicles (the red car and the yellow bicycle). What is the result?"

- **Hop 1** (L1 -> sign_1): Read number from "traffic sign in top-right corner" -> speed_limit = 30
- **Hop 2** (L2 -> sign_1, car_1, bicycle_1): Compare distances: "red car in center" vs "yellow bicycle near bottom" to sign -> closest = car_1
- **Hop 3** (L2 -> car_1, person_1, tree_1): Compare distances: car's center to "person in blue shirt on left" vs "large tree in right background" -> closer to person_1
- **Hop 4** (L2 -> person_1, tree_1, car_1, bicycle_1): Count objects on line between "person in blue" and "tree in right" -> count = 2 (car and bicycle)
- **Hop 5** (L1+L2 -> car_1, bicycle_1): Count wheels: "red car" (4) + "yellow bicycle" (2) -> wheels = 6
- **Final**: 30 x 2 - 6 = 54

**Why this is unambiguous:**
- Every instance has multiple distinguishing features (color + position + distinctive attributes)
- "Red car in center" - unique by color and position
- "Yellow bicycle near bottom" - unique by color and position
- "Person wearing blue shirt on left" - unique by clothing and position
- "Traffic sign in top-right corner" - unique by function and position
- "Large tree with green leaves in right background" - unique by size, color, position

Instance Chain: sign_1 -> car_1 -> person_1/tree_1 -> car_1,bicycle_1 -> final
involved_objects: ["person_1", "car_1", "bicycle_1", "sign_1", "tree_1"] [OK] ALL 5!

---

**GOOD Example (4 hops, instance chain, all objects):**

Query: "Find the object that is furthest from the person. Then identify all objects that share at least one visual attribute (color, shape, or size category) with it. Among those objects, find the one closest to the bicycle. Read any text or number visible on or near that object. If no text exists, use the count of objects to its left. What is this value?"

- **Hop 1** (L2 -> person_1, all others): Find furthest object from person_1 -> sign_1
- **Hop 2** (L2 -> sign_1, car_1, bicycle_1, tree_1): Find objects sharing attributes with sign_1 -> result set contains tree_1 (both have vertical structure)
- **Hop 3** (L2 -> tree_1, bicycle_1): Find which object in result set is closest to bicycle_1 -> tree_1
- **Hop 4** (L1 -> tree_1): Check for text on tree_1, if none, count objects to its left -> 2

Instance Chain: person_1 -> sign_1 -> tree_1 -> bicycle_1 -> tree_1 -> final
involved_objects: ["person_1", "car_1", "bicycle_1", "sign_1", "tree_1"] [OK] ALL 5!

---

**BAD Example (too few hops, too simple):**
- Query: "What color is the car?"
- Problems: Only 1 hop, only 1 object, trivial perception

**BAD Example (few objects, weak chain):**
- Query: "Read the number on the sign and add it to the count of people."
- Problems: Only 2 objects involved, only 2 hops, no instance chain

**BAD Example (hops not logically dependent):**
- Query: "Count red objects and count blue objects. What is the sum?"
- Problems: The two counts are independent (can be done in parallel), not a true chain

**BAD Example (AMBIGUOUS instance references - CRITICAL FAILURE):**
- Query: "Find the car, then count objects to its right."
- Problems:
  - "the car" - which car? If there are 2+ cars, this is ambiguous
  - "to its right" - to the right of what point? Edge? Center? Entire bounding region?
  - Cannot be solved deterministically without borders

**BAD Example (SUBJECTIVE/NON-DETERMINISTIC):**
- Query: "Find the most important object, then count nearby objects."
- Problems:
  - "most important" - subjective, not measurable
  - "nearby" - no clear distance threshold defined
  - Two people would give different answers

**GOOD Fix (making it unambiguous and deterministic):**
- Query: "Find the red sedan car located in the center-bottom area of the image (the only car with visible license plate), then count all objects whose center points are within a horizontal distance of 100 pixels to the right of this car's rightmost edge."
- Why it works:
  - "Red sedan car in center-bottom with visible license plate" - multiple features ensure uniqueness
  - "Center points" - clear reference point
  - "Horizontal distance of 100 pixels to the right of rightmost edge" - precise, measurable criterion
  - Completely deterministic

#### **Final Checklist Before Outputting**

For EACH of your {num_queries_word} sub-queries, verify:
- [ ] **Hop Count:** Does it have {target_hop_count_info}?
- [ ] **Instance Chain:** Does the reasoning hop from instance to instance (A->B->C->...)?
- [ ] **Full Coverage:** Does `involved_objects` contain ALL instances from the combination?
- [ ] **Logical Dependency:** Is the result of each hop REQUIRED for the next hop? (no parallel/skippable hops)
- [ ] **Perception Difficulty:** Does each hop require non-trivial perception (not just "identify object type")?
- [ ] **Logic Complexity:** Does the chain include conditionals, set operations, or multi-step computations?
- [ ] **Numerical Answer:** Is the final answer a specific, computable number?
- [ ] **No Border References:** Does it avoid mentioning detection borders or visualization colors?
- [ ] **[VERY IMPORTANT] Unambiguous Instance References (CRITICAL):** For EACH instance reference in the query, verify:
  - Can someone looking ONLY at the original image (no borders) uniquely identify this exact instance?
  - Would ANY other object in the image match this description? If yes, add more distinguishing details.
  - Does the description use multiple discriminating features (spatial + visual + contextual)?
- [ ] **[VERY IMPORTANT] Deterministic Solution Process (CRITICAL):** Verify:
  - Would two different people arrive at the EXACT SAME answer?
  - Are all spatial relationships clearly defined? (e.g., "left of" = center is to the left)
  - Are all comparisons objective? (e.g., "larger" = larger area, not importance)
  - Are all counting criteria explicit? (what counts as "between", "near", etc.)
  - Is there NO room for subjective interpretation?

**Quality Scoring (aim for 10+):**
- 1 point per hop
- 1 point if ALL instances involved
- 1 point if instance chain covers 4+ different instances
- 1 point if includes conditional logic (if-then)
- 1 point if includes numerical computation beyond counting
- **2 points if ALL instance references are unambiguous and solution is deterministic (CRITICAL)**

**Please Note:** 
- Your task is to generate the JSON content that conforms to the format and requirements above.
- **CRITICAL REMINDERS:** 
  - Each `query` field must be answerable using ONLY the original image (Image 1) without any detection borders or markers.
  - **EVERY query MUST be multi-hop with {target_hop_count_info}.**
  - **EVERY query MUST involve ALL or MOST of the instances in the provided combination list.**
  - The `involved_objects` field must accurately list all instances referenced in the query.
\end{lstlisting}
\end{tcolorbox}

\newpage
\section{Full Prompt for Image Filtering}
\label{app:image-select-prompt}

We also provide the complete prompt used to screen images before multi-hop query synthesis.

\medskip
\noindent
\begin{tcolorbox}[
  breakable,
  title={\textbf{Image Filtering Prompt (Full)}},
  fonttitle=\bfseries,
  colback=white,
  colframe=black,
  boxrule=0.5pt,
  left=4pt,
  right=4pt,
  top=6pt,
  bottom=6pt,
]
\lstset{
  basicstyle=\small\ttfamily,
  breaklines=true,
  columns=flexible,
  frame=none,
  numbers=none,
  showstringspaces=false,
  keepspaces=true,
  breakatwhitespace=false,
  xleftmargin=0pt,
  aboveskip=0pt,
  belowskip=0pt,
}
\begin{lstlisting}
# Role and Goal

You are a professional AI Image Analyst. Your primary goal is to evaluate the complexity of a given image from the perspective of a standard computer vision model. This analysis will identify images that pose a challenge to AI perception, especially in tasks like object detection, counting, and recognition.

# Core Task

For the image I provide, you must:

1.  Evaluate its overall perceptual complexity on a scale of **1 to 10**.
2.  Assess the image quality (High / Medium / Low).
3.  Identify the specific objects or areas that contribute to the complexity.
4.  Output the results as a **single JSON object** as required, with no explanatory text outside of the JSON.

# Key Definitions

## 1. Image Complexity Factors (What makes an image complex for AI?)

*   **Occlusion**: Objects are partially hidden by other objects. The more a key object is occluded, the higher the complexity.
*   **Object Count & Density**: The image contains a large number of distinct objects, especially of the same category (e.g., a crowd of people, a fleet of cars), making counting and individual identification difficult.
*   **Unusual Pose/Angle**: Objects are shown from unconventional viewpoints (e.g., a person doing a handstand, a car seen directly from above).
*   **Complex Interaction**: Objects interact in a way that blurs their individual boundaries (e.g., people hugging, tangled wires, a pile of clothes).
*   **Fine-grained Recognition**: Requires distinguishing between very similar sub-categories (e.g., telling two different bird species apart, or different models of the same car brand).
*   **Challenging Lighting/Shadows**: Harsh shadows, reflections, or poor lighting obscure the shape and details of objects.

## 2. Image Quality Guidelines (How to handle low-quality or unusable images)

Your analysis must focus on clear, perceptible objects.

*   **High/Medium Quality**: The image is generally clear, and its main subjects are recognizable, even if the scene is complex.
*   **Low Quality**: An image should be rated "Low Quality" when it is unusable for precise perceptual analysis due to any of the following reasons:
    *   **Technical Flaws:** The image is too blurry, noisy, severely overexposed, or underexposed, making objects unrecognizable.
    *   **Annotation Impracticality:** **Even if the image itself is clear, the extreme number and density of objects make meaningful, verifiable annotation impossible (e.g., it's impossible to draw a bounding box for or accurately count every single object). Such images lack a "Ground Truth" for model evaluation.**

*   **Crucial**:
    *   **Case 1 (Technical Flaws):** If a scene is chaotic due to blur or lighting issues, making it impossible to distinguish individual objects, its quality should be rated **"Low"** with a **low complexity score (1-3)**.
    *   **Case 2 (Annotation Impracticality):** **If a scene contains a vast, uncountable collection of objects (e.g., a massive but clear crowd where individuals cannot be boxed or counted; a bookshelf or rack packed with countless small items), its quality should also be rated "Low," even if the image is technically clear.**
    *   **Conclusion:** **For both of the "Low Quality" cases described above, the perceptual complexity score should be low (1-3)**, as they are not good test cases for perception but rather issues of data usability or quality itself.

# Output Format Requirements

The output must be a structured **JSON object** (with no additional text). The structure is as follows:

```json
{
  "overall_complexity_score": <Integer, 1-10>,
  "overall_quality_rating": "<High | Medium | Low>",
  "complexity_analysis": "<String, a brief explanation for the score, referencing the complexity factors above>",
  "complex_objects": [
    {
      "object_name": "<Description of the specific object>",
      "generalized_name": "<General category like Person, Animal, Vehicle, Plant, etc.>",
      "reason_for_complexity": [
        "<Select from the list of complexity factors, e.g., 'Occlusion', 'Fine-grained Recognition'>"
      ]
    }
  ]
}
```

# Final Instruction

After I provide an image, strictly follow all the rules above, perform the analysis based on the actual image content, and output a JSON object that strictly adheres to the specified format.
\end{lstlisting}
\end{tcolorbox}

\end{document}